\documentclass[10pt,journal,letterpaper,compsoc]{IEEEtran}

\usepackage[nocompress]{cite}
\usepackage[pdftex]{graphicx}
\graphicspath{{./figures/}}
\DeclareGraphicsExtensions{.pdf}
 \usepackage[cmex10]{amsmath}
\usepackage{array}
\usepackage{algorithmic}
\usepackage{mdwmath}
\usepackage{mdwtab}
\usepackage{eqparbox}
\usepackage[caption=false,font=size,labelfont=sf,textfont=sf]{subfig}
\usepackage{bm}
\usepackage{mdwtab}

\newcommand{ \argmin}[1]{ \underset{#1}{ \operatornamewithlimits{arg \,min}}}
\newcommand{\bt}{\bm{\theta}}
\newcommand{\bT}{\bm{\Theta}}
\newcommand{\bPhi}{\bm{\Phi}}

\newcommand{\bTh}{\widehat{\bT}}

\newcommand{\D}{\vec{\mathrm{d}}}
\newcommand{\KL}{\mathrm{D_{KL}}}

\newcommand{\E}{\mathrm{E}}

\newcommand{\Dir}{\mathrm{Dir}}
\newcommand{\cS}{\mathcal{S}}
\newcommand{\cA}{\mathcal{A}}
\newcommand{\ra}{,}
\newcommand{\EdKL}{\mathrm{PIG}} 

\newcommand{\B}{\mathrm{B}}

\newcommand{\rntwks}{Dense Worlds}
\newcommand{\sntwk}{1-2-3 Worlds}
\newcommand{\sntwks}{1-2-3 Worlds}
\newcommand{\IM}{\mathrm{I_M}}
\newcommand{\IG}{\mathrm{I_G}}
\newcommand{\PIG}{\mathrm{PIG}}
\newcommand{\PEIG}{\mathrm{PEIG}}
\newcommand{\PMC}{\mathrm{PMC}}
\newcommand{\PLC}{\mathrm{PLC}}
\newcommand{\U}{\mathrm{Unemb.}}
\newcommand{\G}{\mathrm{greedy}}
\newcommand{\VI}{\mathrm{VI}}
\newcommand{\MI}{\mathrm{MI}}

\newtheorem{theorem}{Theorem}

\begin{document}
\title{Learning in embodied action-perception loops through exploration}%
\author{Daniel~Y.~Little and~Friedrich~T.~Sommer%
\IEEEcompsocitemizethanks{\IEEEcompsocthanksitem D.Y. Little is with the Department of Molecular and Cellular Biology, University of California at Berkeley, Berkeley, CA 94720.\protect\\
E-mail: dylittle@berkeley.edu
\IEEEcompsocthanksitem F.T. Sommer is with the Redwood Center for Theoretical Neuroscience, University of California, Berkeley, CA 94720\protect\\
E-mail: fsommer@berkeley.edu}%
\thanks{}
}
\IEEEcompsoctitleabstractindextext{%
\begin{abstract}
Although exploratory behaviors are ubiquitous in the animal kingdom, their computational underpinnings are still largely unknown. Behavioral Psychology has identified learning as a primary drive underlying many exploratory behaviors. Exploration is seen as a means for an animal to gather sensory data useful for reducing its ignorance about the environment. While related problems have been addressed in Data Mining and Reinforcement Learning, the computational modeling of learning-driven exploration by embodied agents is largely unrepresented.

Here, we propose a computational theory for learning-driven exploration based on the concept of missing information that allows an agent to identify informative actions using Bayesian inference. We demonstrate that when embodiment constraints are high, agents must actively coordinate their actions to learn efficiently. Compared to earlier approaches, our exploration policy yields more efficient learning across a range of worlds with diverse structures. The improved learning in turn affords greater success in general tasks including navigation and reward gathering. We conclude by discussing how the proposed theory relates to previous information-theoretic objectives of behavior, such as predictive information and the free energy principle, and how it might contribute to a general theory of exploratory behavior.
\end{abstract}
}
\maketitle

 \section{Introduction}
\IEEEPARstart{E}{xploratory} behaviors have been observed and studied in diverse species across the animal kingdom. As one example, approach and investigation of novel stimuli have been studied in vertebrates ranging from fish to birds, reptiles, and mammals \cite{wunschmann1963quantitative, glickman1966curiosity,mikheev1993two,  matsuzawa2006cognitive, stowe2006novel, shinskey2010something}. As another, open field and maze experimental paradigms for studying locomotive exploration in mice and rats have recently been adapted to behavioral studies in zebrafish \cite{stewart2011modeling, stewart2011phenotyping}. Indeed, exploratory behaviors have even been described across a range of invertebrates \cite{holmes1905selection, pisula2009curiosity, byers1998animal,bhl22754}. The prevalence of exploratory behaviors across animal species suggests a fundamental evolutionary advantage, largely believed to derive from the utility of information acquired through such behaviors \cite{kaplan1983cognition,pisula2003costs,pisula2008play,Renner_1988,renner1990neglected}.

Computational models of exploratory behavior, developed predominantly in the field of Reinforcement Learning (RL), have largely focused on the role of exploration in the acquisition of external rewards  \cite{sutton1985learning,sutton1998,kaelbling1996reinforcement,abbeel2005exploration,kawato2007efficient,thrun1992,barto1990computational}. An agent that strictly maximizes the acquisition of known rewards might fall short in finding new, previously unknown, sources of reward. Reward maximization therefore requires balancing between directed harvesting of known rewards (exploitation) and the search for new rewards (exploration) \cite{sato1988learning,kaelbling1996reinforcement,abbeel2005exploration,kawato2007efficient,thrun1992,barto1990computational}. The emphasis in the RL literature on reward acquisition, however, stands in contrast to the dominant psychological theories of exploration. Quoting D. E. Berlyne, a pioneer in the psychology of exploration:
\begin{quote}
As knowledge accumulated about the conditions that govern exploratory behavior and about how quickly it appears after birth, it seemed less and less likely that this behavior could be derivative of hunger, thirst, sexual appetite, pain, fear of pain, and the like, or that stimuli sought through exploration are welcomed because they have previously accompanied satisfaction of these drives. \cite{berlyne1966curiosity}
\end{quote}
Berlyne further suggested ``the most acute motivational problems . . . are those in which the perceptual and intellectual activities are engaged in for their own sake and not simply as aids to handle practical problems'' \cite{berlyne1960conflict}. In fact, a consensus has emerged in behavioral psychology that learning represents the primary drive of exploratory behaviors \cite{loewenstein1994psychology,archer1983exploration,pisula2009curiosity,silvia2005interesting}. To address this gap between computational modeling and behavioral psychology, we introduce here a mathematical framework for studying how behavior effects learning and develop a novel model of learning-driven exploration.

In Computational Neuroscience,  machine learning techniques have been successfully applied towards modeling how the brain might learn the structure underlying sensory signals, e.g., \cite{olshausen1996emergence,rehn2007network,lewicki2002efficient,5766096,1556155,serre2007robust,crutchfield1987equations}. Generally, these methods focus on passive learning where the learning system can not directly effect the sensory input it receives. Exploration, in contrast, is inherently active, and can only occur in the context of a closed-action perception loop. Learning in closed action-perception loops differs from passive learning in two important aspects \cite{gordon2011toward}. First, a learning agent's internal model of the world must keep track of how actions change the sensory input. Sensorimotor contingencies, such as the way visual scenes change as we shift our gaze or move our head, must be taken into account to properly attribute changes in sensory signals to their causes. This is perhaps reflected in neuroanatomy where tight sensory-motor integration has been reported at all levels of the brain \cite{pmid20696186,pmid16226588}. Though often taken for granted, sensor-motor contingencies must actually be learned during the course of development as is eloquently expressed in the explorative behaviors of young infants (e.g., grasping and manipulating objects during proprioceptive exploration or bringing them into visual view during intermodal exploration) \cite{rochat1989object, o2001sensorimotor,noe2004action}. 

The second crucial aspect of learning in a closed action-perception loop is that actions direct the acquisition of sensory data. To discover what is inside an unfamiliar box, a curious child must open it. To learn about the world, scientists perform experiments. Directing the acquisition of data is particularly important for embodied agents whose actuators and sensors are physically confined. Since the most informative data may not always be accessible to a physical sensor, embodiment may constrain an exploring agent and require that it coordinates its actions to retrieve useful data.

In the model we propose here, an agent moving between discrete states in a world has to learn how its actions influence its state transitions. The underlying transition dynamics are governed by a Controllable Markov Chain (CMC). Within this simple framework, various utility functions for guiding exploratory behaviors will be studied, as well as several methods for coordinating actions over time. The different exploratory strategies are compared in their rate of learning. 
\section{Model}

\begin{table*}
\centering
\caption{Table of Information Measures}
\label{table1}
\begin{IEEEeqnarraybox}[\IEEEeqnarraystrutmode \IEEEeqnarraystrutsizeadd{2pt}{1pt}]%
{c/v/c/v/c}
\IEEEeqnarrayrulerow\\
\mbox{Name used here, Abbreviation (Equation Number)}&&\mbox{Name used in [Reference]}&&\mbox{Mathematical expression}\\
\IEEEeqnarraydblrulerow\\
\IEEEeqnarrayseprow[5pt]\\
Missing\;Information,\;\IM\;(\ref{eq:IM})&& Missing\;Information\;\mbox{\cite{pfaffelhuber1972}} && \sum_{s\in \cS, a\in \cA}\KL(\bT_{s,a\ra :}||\bTh_{s,a\ra :}) \\
\IEEEeqnarrayseprow[10pt]\\
Information\;Gain,\;\IG\;(\ref{eq:IG})&& \; && \IM(\bTh) - \IM(\bTh^{a,s,s^*})\\
\IEEEeqnarrayseprow[10pt]\\
Predicted\;Information\;Gain,\;PIG\;(\ref{eq:PIG})&& Information\;Gain\;\mbox{\cite{nelson2005}} && E_{s^*|a,s,\D_{a,s}}\left[\KL(\bTh_{a,s\ra:}||\bTh^{a,s,s^*}_{a,s\ra:})\right]\\
\IEEEeqnarrayseprow[10pt]\\
Posterior\;Expected\;Information\;Gain,\;PEIG\;(\ref{eq:PEIG})&& KL-Divergence\;\mbox{\cite{storck1995}}\;or\;Surprise\;\mbox{\cite{itti2006}} && \sum_{s\in \cS, a \in \cA}\KL(\bTh^{past}_{a,s\ra :}\parallel \bTh^{current}_{a,s\ra :})\\
\IEEEeqnarrayseprow[10pt]\\
Predicted\;Mode\;Change,\;PMC\;(\ref{eq:PMC})&& Probability\;Gain\;\mbox{\cite{nelson2005}} && \sum_{s^*}\bTh_{a,s,s^*}\left[\max_{s'}\bTh^{a,s,s^*}_{a,s,s'}-\max_{s'}\bTh_{a,s,s'}\right]\\
\IEEEeqnarrayseprow[10pt]\\
Predicted\;L_1\;Change,\;PLC\;(\ref{eq:PLC})&& Impact\;\mbox{\cite{nelson2005}} && \sum_{s^*}\bTh_{s,a\ra s^*}\left[\frac{1}{N}\sum_{s'}\left|\bTh^{a,s,s^*}_{a,s\ra s'}-\bTh_{a,s\ra s'}\right|\right]\\
\IEEEeqnarraystrutsize{0pt}{0pt}\\
\IEEEeqnarrayseprow[5pt]\\
\IEEEeqnarrayrulerow
\end{IEEEeqnarraybox}
\end{table*}

\subsection{Mathematical framework for embodied active learning}
{\it Controllable Markov chains (CMCs)} are a simple extension of Markov chains that incorporate a control variable for switching between different transition distributions \cite{gimbert2007pure}. Formally, a CMC is a 3-tuple $(\cS, \cA, \bT)$ where:
\begin{itemize}
\item $\cS$ is a finite set of the possible {\it states} of the system (for example, the possible location of an agent in its world). $N=|\cS|$

\item $\cA$ is a finite set of the possible control values, i.e., the {\it actions} the agent can choose. $M=|\cA|$

\item $\bT$ is a 3-dimensional {\it CMC kernel} describing the probability of transitions between states given an action (for example, the probability an agent moves from state s to state s' when it chooses action a):
\begin{IEEEeqnarray*}{l}
p_{a,s}(s'| \bT)=\bT_{a,s\ra s'}\label{eq:cd}\IEEEyesnumber\\
\sum_{s'}\bT_{a,s\ra s'}=1%
\end{IEEEeqnarray*}
\end{itemize}

For any fixed action, $\bT_{a,:\ra :}$ is a (two-dimensional) stochastic matrix describing a Markov process. Each column in this matrix $\bT_{a,s\ra :}$ defines a {\it transition distribution} which is a categorical (finite discrete) distribution specifying the likelihoods for the next state.  

The CMC provides a simple mathematical framework for modeling exploration in embodied action-perception loops. At every time step, an exploring agent is allowed to freely select any action $a \in \cA$. The learning task of the exploring agent is to build from observed transitions an estimate, the {\it internal model} $\bTh$, of the true CMC kernel,  the {\it world} $\bT$. We assume that the explorer begins with limited information about the world in the form of a prior and must improve its estimate by acting in the world and gathering observations. The states can be directly observed by the agent, i.e. the system is not hidden. In the CMC framework, an agent's immediate ability to interact with and observe the world is limited by the current state. This restriction models the {\it embodiment} of the agent. To ameliorate the myopia imparted by its embodiment, an agent can coordinate its actions over time. Our primary question is how action policies can optimize the speed and efficiency of learning in embodied action-perception loops.

\subsection{Information-theoretic assessment of learning}
To assess an agent's success towards learning, we determine the {\it missing information} $\IM$ in its internal model (as proposed by Pfaffelhuber \cite{pfaffelhuber1972}). We do this by first calculating the Kullback-Leibler (KL) divergence of the internal model from the world for each transition distribution:
\begin{equation*}
\KL(\bT_{a,s\ra :}\parallel\bTh_{s,a\ra :}):=\sum_{s'=1}^N\bT_{s,a\ra s'}\log_2\left(\frac{\bT_{s,a\ra s'}}{\bTh_{s,a\ra s'}}\right)
\end{equation*}
The KL-divergence can be interpreted as the expected amount of information in bits, lost when observations (following the true distribution) are communicated using an encoding scheme optimized for the estimated distribution \cite{cover1991elements}. The loss is large when the two distributions differ greatly and zero when they are identical. The missing information is then calculated as the sum of the KL-divergences for each transition distribution:
\begin{equation}
\IM(\bT\parallel\bTh):=\sum_{s\in \cS, a\in \cA}\KL(\bT_{s,a\ra :}||\bTh_{s,a\ra :})
\label{eq:IM}
\end{equation}
We will use missing information (\ref{eq:IM}) to assess learning and to compare the performance of different explorative strategies. Steeper decreases in missing information over time represent faster learning and thus more efficient explorative strategies. Table 1 has been included as a reference guide for the various measures discussed in this manuscript for assessing or guiding exploration.

\subsection{Bayesian inference learning in an agent}
\label{ss:bilita}
During exploration, a learning agent will gather data in the form of experienced state transitions. At every time step, it can update its internal model $\bTh$ with its last observation. 

Assuming the transition probabilities are drawn from a {\it prior distribution} $f$ (a distribution over CMC kernels), and letting $\D$ be the history of observed transitions, the Bayesian estimate $\bTh_{s,a\ra s'}$ for the probability of transitioning to state $s'$ from state $s$ under action $a$ is given by:
\begin{IEEEeqnarray*}{rCl}
\bTh_{a,s\ra s'}:&=&p_{a,s}(s'|\D)\\
&=&\int_{\bT} p_{a,s}(s',\bT|\D)d\bT\\
&=&\int_{\bT}p_{a,s}(s'|\bT,\D)f(\bT|\D)d\bT\\
&=&\int_{\bT}p_{a,s}(s'| \bT)f(\bT|\D)d \bT\\
&=&\int_{\bT}\bT_{a,s\ra s'}f(\bT|\D)d \bT\\
&=&\E_{\bT}[\bT_{a,s\ra s'}|\D]:=\E_{\bT|\D}[\bT_{a,s\ra s'}]\IEEEyesnumber %
\label{eq:be}
\end{IEEEeqnarray*}
For discrete priors the above integrals would be replaced with summations. At the beginning of exploration, when the history $\D$ is empty, $f(\bT|\D)$ is simply the prior $f(\bT)$. Equation (\ref{eq:be}) demonstrates that the Bayesian estimate is equivalent to the expected value of the true CMC kernel given the data. If we further assume that each transition distribution $\bT_{a,s,:}$ is independently drawn from the marginals $f_{a,s}$ of the distribution $f$ and we let $\D_{a,s}$ be the history of state transitions experienced when taking action $a$ in state $s$, (\ref{eq:be}) simplifies to the independent estimation of each transition distribution:
\begin{IEEEeqnarray*}{rCl}
\bTh_{a,s\ra s'}&=&\int_{\bT}\bT_{a,s\ra s'}f(\bT|\D)d \bT\\
&=&\int_{\bT_{a,s\ra :}}\bT_{a,s\ra s'}f_{a,s}(\bT_{a,s\ra :}|\D_{a,s})d \bT_{a,s\ra :}\\
&=&\E_{\bT_{a,s,:}|\D_{a,s}}[\bT_{a,s\ra s'}] \IEEEyesnumber %
\label{eq:be2}
\end{IEEEeqnarray*}
In the following theorem, we demonstrate that the Bayesian estimator minimizes the expected missing information and thus is the best estimate under our objective function (\ref{eq:IM}).

\begin{theorem}
Let $\bT$ and $\Phi$ be CMC kernels. $\bT$ describes the ground truth environment generated from a prior distribution. $\Phi$ is any internal model of the agent. Then the expected missing information between $\bT$ and an internal model $\Phi$, given data $\D$, is minimized by the Bayesian estimate $\bTh$:
\begin{equation}
\bTh=\argmin{\bPhi}\;E_{\bT|\D}\left[\IM(\bT\parallel\bPhi)\right] \\
\end{equation}
\end{theorem}
\begin{IEEEproof}
Since missing information is simply the sum of the KL-divergence for each transition kernel (\ref{eq:IM}), minimizing missing information is equivalent to independently minimizing these KL-divergences:
\begin{IEEEeqnarray*}{l}
\argmin{\bPhi_{a,s\ra :}}\;E_{\bT_{a,s\ra :}|\D}\left[\KL\left(\bT_{a,s\ra :}\parallel \bPhi_{a,s\ra :}\right)\right]\\
\;\;= \argmin{\bPhi_{a,s\ra :}}\;E_{\bT_{a,s\ra :}|\D}\left[ \sum_{s'} \bT_{a,s\ra s'} \log_2\left(\frac{\bT_{a,s\ra s'}}{\bPhi_{a,s\ra s'}}\right)\right] \\
\;\;= \argmin{\bPhi_{a,s\ra :}}\;E_{\bT_{a,s\ra :}|\D}\left[ \sum_{s'} \bT_{a,s\ra s'} \log_2 \bT_{a,s\ra s'}\right]\\
\;\;\;\;\;\;\;\;\;\;\;\;\;\;\;\;\;\;\;\;\;\;\; - E_{\bT_{a,s\ra :}|\D}\left[ \sum_{s'} \bT_{a,s\ra s'} \log_2 \bPhi_{a,s\ra s'}\right] \\
\;\;= \argmin{\bPhi_{a,s\ra :}}\;- E_{\bT_{a,s\ra :}|\D}\left[ \sum_{s'} \bT_{a,s\ra s'} \log_2 \bPhi_{a,s\ra s'}\right] \\
\;\;=  \argmin{\bPhi_{a,s\ra :}}\;-  \sum_{s'} E_{\bT_{a,s\ra :}|\D}\left[\bT_{a,s\ra s'}\right] \log_2 \bPhi_{a,s\ra s'} \\
\;\;=  \argmin{\bPhi_{a,s\ra :}}\;H\left[E_{\bT_{a,s\ra :}|\D}\left[\bT_{a,s\ra :}\right];\bPhi_{a,s\ra :}\right]%
\end{IEEEeqnarray*}
Here $H[\theta;\phi]$ denotes the cross-entropy \cite{cover1991elements}. Then, by Gibb's inequality \cite{cover1991elements} we conclude:
\begin{IEEEeqnarray*}{l}
\argmin{\bPhi_{a,s\ra :}}\;H\left[E_{\bT_{a,s\ra :}|\D}\left[\bT_{a,s\ra :}\right];\bPhi_{a,s\ra :}\right]\\
\;\;= E_{\bT_{a,s\ra :}|\D}\left[\bT_{a,s\ra :}\right]\\
\;\;=\bTh_{a,s\ra :}(\D)%
\end{IEEEeqnarray*}
\end{IEEEproof}

The analytical form for the Bayesian estimate will depend on the prior. In the following section, we introduce three classes of CMCs which will be considered in this study and specify the Bayesian estimates for each.

\begin{figure}[!t]
\centering
\includegraphics[scale=0.7]{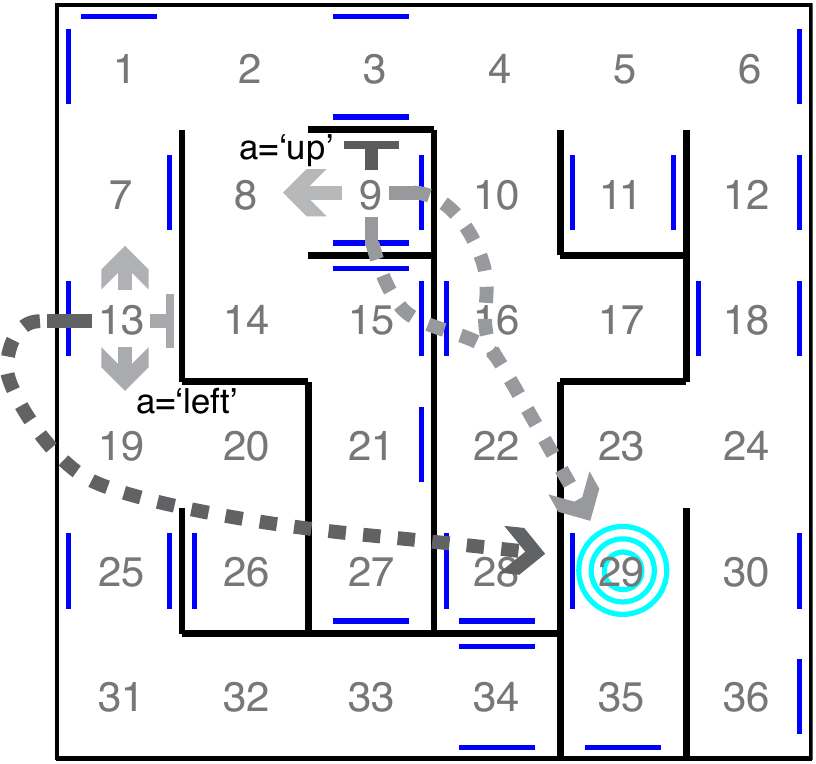}
\caption{Example maze. The 36 states correspond to rooms in a maze. The 4 actions correspond to noisy translations in the cardinal directions. Two transition distributions are depicted, each by a set of 4 arrows emanating from their starting state. Flat-headed arrows represent translations into walls, resulting in staying in the same room. Dashed arrows represent translation into a portal (blue lines) leading to the base state (blue target). The shading of an arrow indicates the probability of the transition (darker color represents higher probability).}
\label{fig:maze}
\end{figure}

\subsection{Three test environments for studying exploration}
\label{ss:worlds}
In the course of exploration, the data an agent accumulates will depend on both its behavioral strategy as well as the world structure. Studying diverse environments, i.e., CMCs that differ greatly in structure, will help us to investigate how world structure effects the relative performance of different exploratory strategies and to identify action policies that produce efficient learning under broad conditions. 

The three different classes of test environments to be investigated will be called Dense Worlds, Mazes, and 1-2-3 Worlds. For each class, random CMCs are generated by drawing the transition distributions from specific distributions. These generative distributions are also given to the agents as priors for performing Bayesian inference.

{\it 1) \rntwks{}} correspond to complete directed probability graphs with $N=10$ states and $M=4$ actions. Each transition distribution is independently drawn from a Dirichlet distribution:
\begin{IEEEeqnarray*}{l}
\bT_{s,a,\ra :} \sim \Dir(1):=\frac{\prod_{s'} {\bT_{s,a, s'}}^{\alpha-1}}{\B(\alpha)}\IEEEyesnumber\\
\;\;\;\;\;\;\;\;\B(\alpha):=\frac{\Gamma(\alpha)^N}{\Gamma(N\alpha)}\\
\;\;\;\;\;\;\;\;\Gamma(\alpha):=\int_0^\infty t^{\alpha-1}e^{-t}\mathrm{d} t%
\end{IEEEeqnarray*}
The Dirichlet distribution is the conjugate prior of the categorical distribution (\ref{eq:cd})  and thus a natural distribution for generating CMCs. It is parametrized by a concentration factor $\alpha$ that determines how much weight in the Dirichlet distribution is centered at the midpoint of the simplex, the space of all possible transition distributions. The midpoint corresponds to the uniform categorical distribution. For \rntwks{}, we use a concentration parameter $\alpha=1$ which results in a uniform distribution over the simplex. An example is depicted graphically in Fig. S1.

The Bayesian estimate for \rntwks{} has the following closed-form expression:
\begin{equation}
\bTh_{a,s\ra s'}=\frac{1+\sum_{k=1}^{K}\delta_{s',\D_{a,s}[k]}}{K+N}
\label{eq:sh} 
\end{equation}
where $K$ is the length of the history $\D_{a,s}$ and $\delta_{x,y}$ is the Kronecker delta (i.e. $\delta_{x,y}$ is 1 if $x=y$ and 0 otherwise). Equation (\ref{eq:sh}) reveals that the Bayesian estimate is simply the relative frequencies of the observed data with the addition of one fictitious count per transition. The incorporation of this fictitious observation is referred to as Laplace smoothing and is often performed to avoid over-fitting \cite{manning2008introduction}. The derivation of Laplace smoothing from Bayesian inference over a Dirichlet prior is a well known result \cite{mackay1995hierarchical}.

{\it 2) Mazes} consist of $N=36$ states corresponding to rooms in a randomly generated 6 by 6 maze and $M=4$ actions corresponding to noisy translations, each biased towards one of the four cardinal directions "up", "down", "left" and "right". An example is depicted in Fig. \ref{fig:maze}. Walking into a wall causes the agent to remain in its current location. There are 30 transporters randomly distributed amongst the walls which lead to a base state. Each maze has a single, randomly chosen base state (concentric rings in Fig. \ref{fig:maze}).  All transitions that do not correspond to a single translation are assigned a probability of zero. The non-zero probabilities are drawn from a Dirichlet distribution with concentration parameter $\alpha=0.25$. The highest probability is assigned to the state corresponding to the cardinal direction of the action. The small concentration parameter distributes more probability weight in the corners of the simplex corresponding to deterministic transitions. This results in Maze transitions have strong biases towards an actions associated cardinal direction.

Agents in  Mazes must estimate the non-zero transitions using the Dirichlet prior without knowledge of each action's assigned cardinal direction. Similar to (\ref{eq:sh}), the Bayesian estimate for maze transitions is given by:  
\begin{equation}
\bTh_{a,s\ra s'}=\frac{0.25+\sum_{k=1}^{K}\delta_{s',\D_{a,s}[k]}}{K+0.25\cdot N_{a,s}}
\label{eq:shm} 
\end{equation}
where $N_{a,s}$ is the number of non-zero probability states in the transition distribution $\bT_{a,s\ra :}$. As with \rntwks{}, the Bayesian estimate (\ref{eq:shm}) for mazes is a Laplace smoothed histogram.

{\it 3) \sntwks{}} consists of $N=20$ states and $M=3$ actions. In a given state, action $a=1$ moves the agent deterministically to a single target state, action $a=2$ brings the agent with probability $0.5$ to one of two possible target states, and action $a=3$ brings the agent with probability $0.333$ to one of 3 potential target states. The target states are randomly and independently selected for each action taken in each state. To create an absorbing state, the probability that state 1 is among the targets of action $a$ is set to $1-0.75^a$. The probability for all other states to be selected as targets is uniform. Explicitly, letting $\Omega_a$ be the set of all admissible transition distributions for action $a$:
\begin{displaymath}
\Omega_a:=\{\bt\in I\!\!R^N|\sum_{s'}\bt_{s'}=1\mathrm{\;and\;} \bt_{s'}\in\{0,\frac{1}{a}\} \forall s'\}\\
\end{displaymath}
the transition distributions are drawn from the following distribution:
\begin{equation}
p(\bT_{a,s\ra :})=\left \{ \begin{array}{cl}
0&\mbox{if $\bT_{a,s\ra :}\notin\Omega_a$}\\
\cfrac{1-0.75^a}{\binom{N-1}{a-1}}&\mbox{else if $\bT_{a,s\ra 1}=\frac{1}{a}$}\\
\\
\cfrac{1-(1-0.75^a)}{\binom{N-1}{a}}&\mbox{otherwise}
\end{array}
\right.
\label{eq:sne}
\end{equation}

If this process results in a non ergodic CMC, it is discarded and a new CMC is generated. A CMC as ergodic if, for every ordered pair of states, there exist an action policy under which an agent starting at the first state will eventually reach the second. An example \sntwk{} is depicted in Fig. S2.

Bayesian inference in \sntwks{} differs greatly from Mazes and \rntwks{} because of its discrete prior.  Essentially, state transitions that have been observed are accurately estimated, while the remaining probability weight is distributed across those states that have not yet been experienced (preferentially to state 1 and uniformly across other states). Explicitly, if $a,s\rightarrow s'$ has been previously observed, then the Bayesian estimate for $\bT_{a,s,s'}$ is given by:
\begin{displaymath}
\bTh_{a,s\ra s'}=\frac{1}{a}
\end{displaymath}
If $a,s\rightarrow s'$ has not been observed but $a,s\rightarrow 1$ has, then the Bayesian estimate is given by:
\begin{IEEEeqnarray*}{l}
\bTh_{a,s\ra s'}=\frac{1-\frac{|\cS^*|}{a}}{N-T}%
\end{IEEEeqnarray*}
Here $T$ is the number of target states that have already been observed:
\begin{IEEEeqnarray*}{l}
T:=|\{s^*\in \D_{a,s}\}|%
\end{IEEEeqnarray*}
Finally, if neither $a,s\rightarrow s'$ nor $a,s\rightarrow 1$ have been observed, then the Bayesian estimate is:
\begin{equation*}
\bTh_{a,s\ra s'}=\left \{ \begin{array}{cl}
\cfrac{1-0.75^{a}}{1+\left(\binom{a-1}{T}-1\right)*0.75^{a}}\cdot \frac{1}{a}&\mbox{if $s'=1$}\\
\\
\cfrac{1-\left(\frac{T}{a}+\bTh_{a,s,1}\right)}{N-T-1} &\mbox{otherwise}
\end{array}
\right.
\end{equation*}

\section{Results}
\begin{figure*}[!t]
\centering
\includegraphics[scale=0.6]{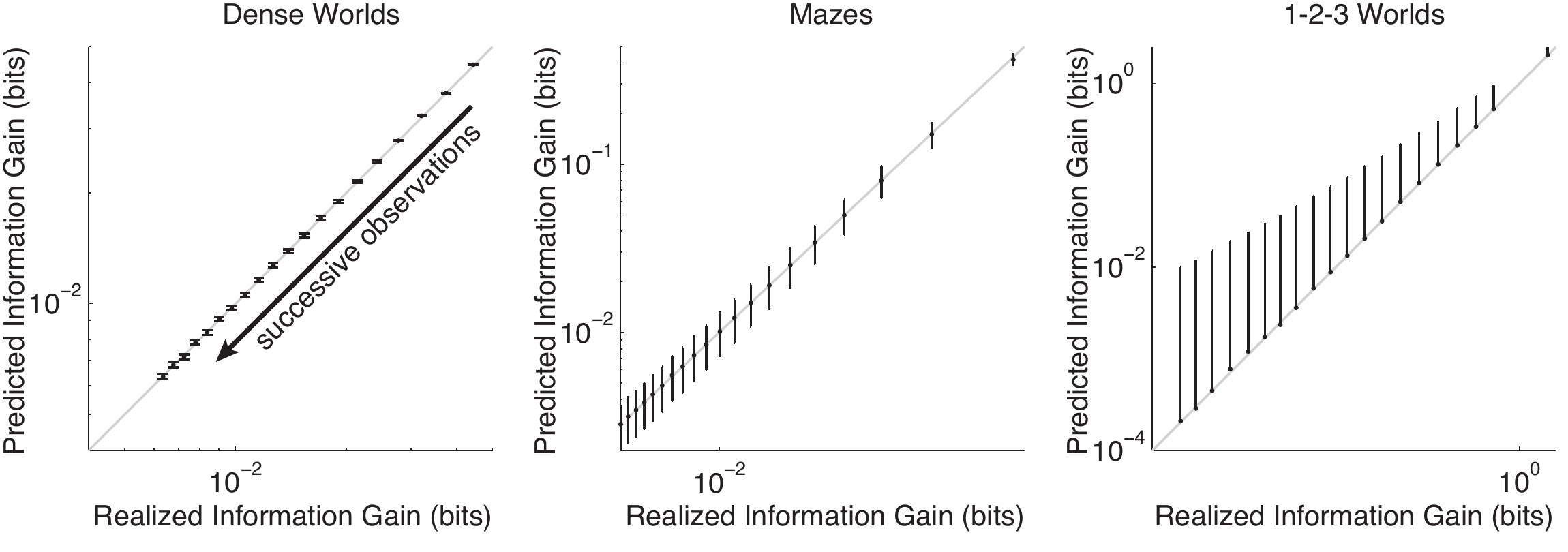}
\caption{Accuracy of predicted information gain. The average predicted information gain is plotted against the average realized information gain. Averages are taken over 200 CMCs, $N\times M$ transition distributions, and 50 trials. Error bars depict standard deviations (only plotted above the mean for \sntwks{}). The arrow indicates the direction of increasing numbers of observations (top-right
=none, bottom-left=19). The unity lines are drawn in gray.}
\label{fig:PIGvsIG}
\end{figure*}

\subsection{Assessing the information-theoretic value of planned actions}
The central question to be addressed is how actions effect the learning process in embodied action-perception loops. Ideally, actions should be chosen so that the missing information (\ref{eq:IM}) decreases as fast as possible. As discussed in Section~\ref{ss:bilita}, the Bayesian estimate minimizes the expected missing information. We will assume that an agent continually updates its internal model accordingly from the observations it receives. The Bayesian estimate, however, does not indicate which action will optimize the utility of future data. Towards this objective, an agent should try to predict the impact a new observation will have on its missing information. We call the decrease in missing information between two internal models the {\it information gain} ($\IG$). Letting $\bTh$ be a current model derived from data $\D$ and $\bTh^{a,s,s^*}$ be an updated model derived from adding an observation of $a,s \rightarrow s^*$ to $\D$, the information gain for this observation is:
\begin{IEEEeqnarray*}{l}
\IG(a,s, s^*):=\IM(\bT\parallel\bTh)-\IM(\bT\parallel\bTh^{a,s, s^*})\\
\;\;\;\;=\KL(\bT_{a,s,:}\parallel\bTh_{a,s,:})-\KL(\bT_{a,s,:}\parallel\bTh^{a,s, s^*}_{a,s,:})\\
\;\;\;\;=\sum_{s'}\bT_{a,s,s'}\log_2\frac{\bT_{a,s,s'}}{\bTh_{a,s,s'}}-\bT_{a,s,s'}\log_2\frac{\bT_{a,s,s'}}{\bTh^{a,s, s^*}_{a,s,s'}}\\
\;\;\;\;=\sum_{s'}\bT_{a,s,s'}\log2\frac{\bTh^{a,s, s^*}_{a,s,s'}}{\bTh_{a,s,s'}}\label{eq:IG}\IEEEyesnumber%
\end{IEEEeqnarray*}
Calculating the information gained from taking action $a$ in state $s$ would therefore require knowing $\bT$ as well as $s^*$. An agent can only infer former and can only know the latter after it has executed the action. In the following theorem, however, we derive a closed-form expression for the expected information gain, which we shall call the {\it predicted information gain} (PIG).
\begin{theorem}
Let $\bT$ be a CMC kernel whose transition distributions are independently generated from prior distributions. If an agent is in state $s$ and has previously collected data $\D$, then the expected information gain for taking action $a$ and observing the resultant state $S^*$ is given by:
\begin{IEEEeqnarray*}{rCl}
\PIG(a,s):&=&\E_{s^*,\bT|\D}[\IG(a,s,s^*)]\\
&=&\sum_{s^*}\bTh_{a,s,s^*}\KL(\bTh_{a,s\ra:}\parallel\bTh^{a,s,s^*}_{a,s\ra:}) \label{eq:PIG}\IEEEyesnumber
\label{eq:kl}%
\end{IEEEeqnarray*}
where $\bTh$ is the current internal model of the agent and $\bTh^{a,s,s^*}$ is what the internal model would become if it were updated with an observation $s^*$ resulting from a prospective new action $a$.  
\end{theorem}
\begin{IEEEproof}
\begin{IEEEeqnarray*}{l}
\E_{s^*,\bT|\D}[\IG(a,s,s^*)]\\
=\E_{s^*,\bT|\D}\left[\sum_{s'}\bT_{a,s,s'}\log2\left(\frac{\bTh^{a,s, s^*}_{a,s,s'}}{\bTh_{a,s,s'}}\right)\right]\\
=\E_{s^*,\bT_{a,s,:}|\D_{a,s}}\left[\sum_{s'}\bT_{a,s,s'}\log2\left(\frac{\bTh^{a,s, s^*}_{a,s,s'}}{\bTh_{a,s,s'}}\right)\right]\\
=\E_{s^*|\D_{a,s}}\left[\E_{\bT|\D_{a,s},s^*}\left[\sum_{s'}\bT_{a,s,s'}\log2\left(\frac{\bTh^{a,s, s^*}_{a,s,s'}}{\bTh_{a,s,s'}}\right)\right]\right]\\
=\E_{s^*|\D_{a,s}}\left[\sum_{s'}\E_{\bT|\D_{a,s},s^*}\left[\bT_{a,s,s'}\right]\log2\left(\frac{\bTh^{a,s, s^*}_{a,s,s'}}{\bTh_{a,s,s'}}\right)\right]\\
=\E_{s^*|\D_{a,s}}\left[\sum_{s'} \bTh^{a,s,s^*}_{a,s\ra s'} \log_2\left(\frac{\bTh^{a,s,s^*}_{a,s\ra s'}}{\bTh_{a,s\ra s'}}\right)\right]\\
=\E_{s^*|\D_{a,s}}\left[\KL(\bTh_{a,s\ra:}\parallel\bTh^{a,s,s^*}_{a,s\ra:})\right]\\
=\sum_{s^*}p(s^*|\D_{a,s})\KL(\bTh_{a,s\ra:}\parallel\bTh^{a,s,s^*}_{a,s\ra:})\\
=\sum_{s^*}\bTh_{a,s,s^*}\KL(\bTh_{a,s\ra:}\parallel\bTh^{a,s,s^*}_{a,s\ra:})%
\end{IEEEeqnarray*}
\end{IEEEproof}

Notice, (\ref{eq:kl}) can be computed from previously collected data alone. For each class of environments, Fig.~\ref{fig:PIGvsIG} compares the average PIG with the average realized information gain as successive observation are drawn from a transition distribution and used to update a Bayesian estimate. In accordance with Theorem 2, in all three environments PIG accurately predicts the average information gain. Thus, theoretically and empirically, PIG represents an accurate estimate of the average gains towards the learning objective functions that an agent can expect to receive for taking a planned action in a particular state.

Interestingly, the equation for computing PIG, RHS of (\ref{eq:kl}), has been previously considered in the field of psychology, where it was applied to describe human behavior during hypothesis testing \cite{oaksford1994rational,nelson2005,klayman1987confirmation}. To our knowledge, however, its equality to the expected decrease in missing information (Theorem 2) has not been previously shown.

\subsection{Control learners: unembodied and random action}

\begin{figure*}[!t]
\centering
\includegraphics[scale=0.7]{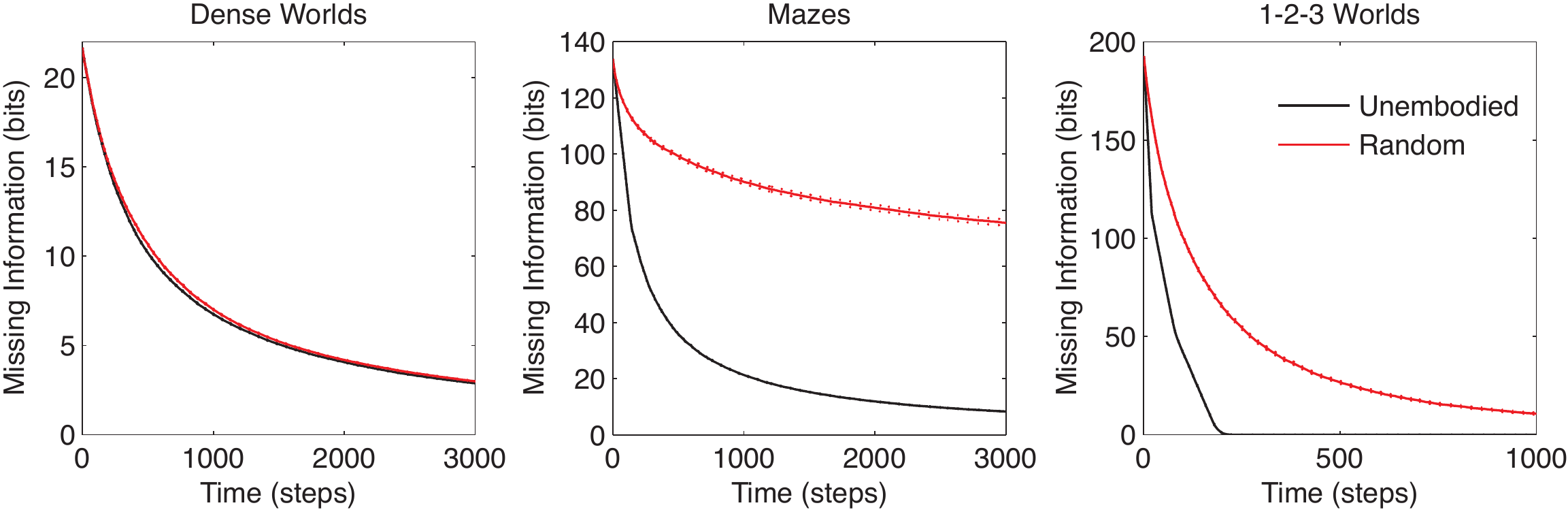}
\caption{Learning curves for control strategies. The average missing information is plotted over exploration time for the unembodied positive control and random action baseline control. Standard errors are plotted as dotted lines above and below learning curves. (n=200)}
\label{fig:Controls}
\end{figure*}

During exploration, an embodied agent can choose its action but is bound to the state that resulted from its last transition. A simple exploratory strategy would be to always select actions uniformly randomly. We will use such a {\it random action} strategy as a baseline control for learning performance representing a naive explorer.

In contrast to embodied agents, one can also consider an {\it unembodied} agent that is allowed to arbitrarily relocate to a new state before taking an action. For an unembodied agent, optimization of learning becomes much simpler as it decomposes into an independent sampling problem \cite{pfaffelhuber1972}. Since the PIG for each transition distribution decreases monotonically over successive observations (Fig. \ref{fig:PIGvsIG}), learning by an unembodied agent can be optimized by always sampling from the state and action pair with the highest PIG. Thus, learning can be optimized in a greedy fashion:   
\begin{equation}
(a,s)_{\U}:=\operatorname*{arg\,max}_{(a,s)} \EdKL(a,s) \label{eq:Unemb}
\end{equation}
The learning curves of the unembodied agent will serve here as a {\it positive control} as it represents an upper bound for the performance of embodied agents.

An initial comparison between random action and the unembodied control highlights a notable difference among the three classes of environments (Fig. \ref{fig:Controls}). Specifically, the performance margin between the two controls is significant in Mazes and \sntwks{} ($p<0.001$), but not in \rntwks{} ($p>0.01$). The significance was assessed by post-hoc analysis of Friedman's test \cite{hochberg1987multiple} comparing the areas under the two learning curves. Despite using a naive strategy, the random actor is essentially reaching maximum performance in \rntwks{}, suggesting that exploration of this environment is fairly easy. The difference in performance between random action and the unembodied control offers an initial insight into the constraints experienced by embodied agents. A directed exploration strategy may help bridge this gap.

\begin{figure*}[!t]
\centering
\includegraphics[scale=0.7]{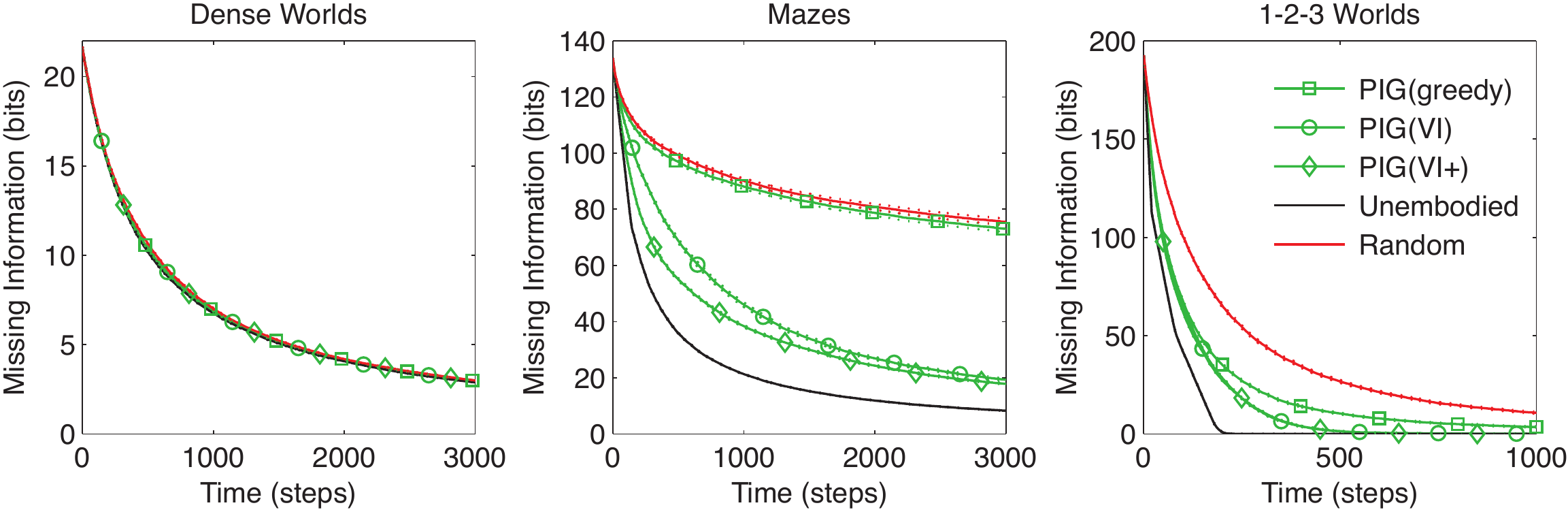}
\caption{Coordinating exploration using predicted information gain. The average missing information is plotted over exploration time for greedy and value-iterated (VI) maximization of PIG. The standard control strategies and the VI+ positive control are also depicted. Standard errors are plotted as dotted lines above and below learning curves. (n=200)}
\label{fig:Coord}
\end{figure*}

\subsection{Exploration strategies based on PIG}
Given that PIG can be computed by an agent using only the data it has already collected (along with its prior), we wondered whether it could be used as a utility function to guide exploration. Since greedy maximization of PIG is optimal for the unembodied agent, one might consider a similar greedy strategy for an embodied agent (PIG(greedy)). The key difference would be that the embodied agent can only select its action but not its current state:
\begin{equation}
a_{\PIG(\G)}:=\operatorname*{arg\,max}_{a} \PIG(a,s) \label{eq:PIG}
\end{equation}
The performance comparison between PIG(greedy) (\ref{eq:PIG}) and the unembodied control (\ref{eq:Unemb}) is of particular interest because the two strategies differ only in that one is embodied but the other is not. Thus differences of their performance reflect the embodiment constraint on learning. As shown in Fig.~\ref{fig:Coord} the performance difference is largest in Maze worlds, moderate though significant in \sntwks{} and smallest in \rntwks{} ($p<0.001$ for Mazes and \sntwks{}, $p>0.001$ for \rntwks{}). To quantify the embodiment constraint faced in a world, we define an {\it embodiment index} as the relative difference between the areas under the learning curves for PIG(greedy) and the unembodied control which average 0.02 for \rntwks{}, 2.59 for Mazes, and 1.27 for \sntwks{}.

Also of particular interest, the comparison between PIG(greedy) and random action provides further insight differentiating the three classes of worlds (Fig.~\ref{fig:Coord}). Whereas PIG(greedy) yielded no improvement over random action in \rntwks{} and Mazes ($p>0.001$), it significantly improved learning in \sntwks{}($p<0.001$), demonstrating that agents benefitted from the information-theoretic utility function only in \sntwks{}. 

Greedy maximization of PIG considers only the immediate gains available and fails to account for the effect an action can have on future utility. In particular, when the potential for information gain is unevenly distributed, it may be necessary to coordinate actions over time to obtain remote but informative observations. Forward estimation of total future PIG over multiple time steps is intractable as the number of action sequences and state outcomes increases exponentially with time. To guide an agent towards maximizing long-term gains of PIG, we instead employ a back-propagation approach previously developed in the field of economics, Value-Iteration (VI) \cite{Bel57}. 
The estimation starts at a distant time point (initialized as $\tau=0$) in the future with initial values equal to the PIG for each state-action pair: 
\begin{equation}
Q_{0}(a,s):=\IG(a,s)\\
\end{equation}
Then propagating backwards in time, we maintain a running total of estimated future value by:
\begin{IEEEeqnarray*}{l}
V_{\tau}(s):=\operatorname*{max}_{a} Q_{\tau}(a,s)\\
Q_{\tau-1}(a,s):=\IG(a,s)+\gamma\sum_{s'\in\cS}\bTh_{s,a\ra s'}\cdot V_{\tau}(s')\IEEEyesnumber\label{eq:vi}%
\end{IEEEeqnarray*}
Here, $0\le\gamma\le1$, is a discount factor reducing the value of gains obtained further in the future. When $\gamma<1$, backward propagation can be continued until convergence. Alternatively, it can simply be executed for a predefined number of steps. Choosing the latter with $\gamma = 1$, we construct a behavioral policy (PIG(VI)) for an agent that coordinates its actions under VI towards maximizing PIG:
\begin{equation}
a_{\PIG(\VI)}:=\operatorname*{arg\,max}_{a} Q_{-10}(a,s);
\end{equation}

Comparing the learning curves in Fig.~\ref{fig:Coord} for PIG(VI) and PIG(greedy) in the three classes of worlds we find that coordination of actions yielded the greatest learning gains in Mazes, with moderate gains also seen in \sntwks{}. In \rntwks{} PIG(VI), like PIG(greedy) and random action, essentially reached maximal learning performance. Along with the results for the embodiment index above, these results support the hypothesis that worlds with high embodiment constraint require agents to coordinate their actions over several time steps to achieve efficient exploration.

Convergence and optimality of the VI algorithm can be guaranteed \cite{Bel57}, but only if the utility function is stationary and the true world structure is known. To assess the impairment resulting from the use of the internal model in VI (\ref{eq:vi}), we constructed a second positive control, PIG(VI+), which is given the true CMC kernel $\bT$ for use during coordinated maximization of PIG under VI. Under this strategy, $\bT$ is used only to coordinate the selection of actions and is not incorporated into the Bayesian estimate or the PIG utility function. Comparing the PIG(VI) agent to the PIG(VI+) control, we find that they only differ in Mazes, and this difference is relatively small compared to the gains made over random or greedy behaviors (Fig. \ref{fig:Coord}). Altogether these results suggest that PIG(VI) may be an effective strategy employable by embodied agents for coordinating explorative actions towards learning.

From the results so far the picture emerges that the three classes of environments offer very different challenges for the exploring agent. \rntwks{} are easy to explore. Mazes require policies that coordinate actions over time but exhibit little sensitivity to the particular choice in utility function. \sntwks{} also require coordination of actions over time, though to a lesser extent than Mazes. Unlike in Mazes, however, agents in \sntwks{} strongly 
benefit from the information-theoretically derived utility function PIG.  

\begin{figure}[!t]
\captionsetup[subfloat]{labelformat=simple}
\centering
\includegraphics[scale=0.9]{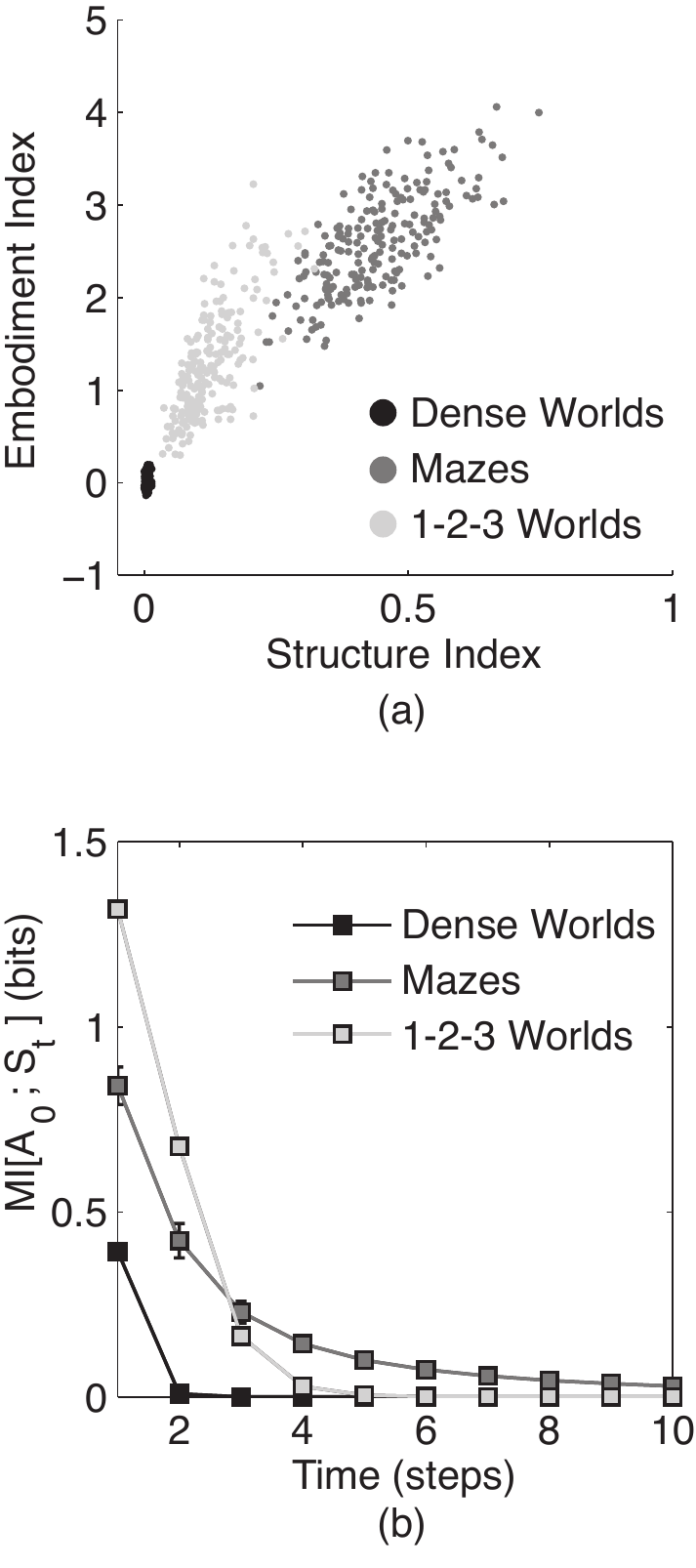}
\caption{Quantifying the structure of the worlds. (a) The embodiment index, defined in Section 3.3, is plotted against the structure index for each of 200 \rntwks{}, Mazes, and \sntwks{}. (b) The average controllability, as measured by the mutual information between an action and a future state, is plotted as a function of the number of time steps the state lies in the future (n=200). The error bars depict standard deviations.}
\label{fig:Structure}
\end{figure}

\subsection{Structural features of the three worlds}

We next asked how structural differences in the three classes of environments correlated with the above differences in exploration performance. In particular we considered two structural features of the worlds, their tendency to draw agents into a biased distribution over states and how tightly an action controls the future states of the agent.    

{\it State bias:} To assess how strongly a world biases the state distribution of agents we consider the equilibrium distribution under an undirected action policy, random action. The equilibrium distribution $\Psi$ is the limit distribution over states after many time steps. To quantify the bias of this distribution, we compute a {\it structure index} (SI) as the relative difference between its entropy $H(\Psi)$ and the entropy of the uniform distribution $H(U)$:
\begin{equation*}
SI(\Psi):=\frac{H(U)-H(\Psi)}{H(U)}
\end{equation*}
where:
\begin{equation*}
H(p(s)):=-\sum_{s\in\cS}p(s)\log_2(p(s))
\end{equation*}
The structure index values for 200 worlds in each class of environment are plotted against the embodiment index (defined in section 3.3) in Fig.~\ref{fig:Structure}a. As depicted, the embodiment index correlates strongly with the structure index. Thus, 
the state bias seems to represent a significant challenge embodied agents face during exploration. 

{\it Controllability:} To measure the capacity for an agent to control its state trajectory we computed the mutual information between a random action and a future state:
\begin{equation*}
\MI[A_0,S_t|s_0]=\sum_{a_0\in\cA, s_t\in\cS} p(a_0,s_t|s_0)\log_2\left(\frac{p(s_t|a_0,s_0)}{p(s_t|s_0)}\right)
\end{equation*}
As shown in Fig. \ref{fig:Structure}b, an action in a Maze or \sntwk{} has significantly more impact on future states than an action in \rntwks{}. Controllability is required for effective coordination of actions, such as under PIG(VI). In Mazes, where actions can significantly effect states far into the future, agents yielded the largest gains from coordinated actions. However, controllability, while necessary, is not sufficient for coordinated actions to have the potential of improving learning. For example, a non-ergodic world might have high controllability but not allow an embodied agent to ever reach a large isolated set of states, regardless of whether it coordinated its actions or not. In such a world, an unembodied agent could reach the isolated states and thereby gain a learning opportunity inaccessible to any embodied agent.

\subsection{Comparison to previous explorative strategies}
\begin{figure*}[!t]
\centering
\includegraphics[scale=0.7]{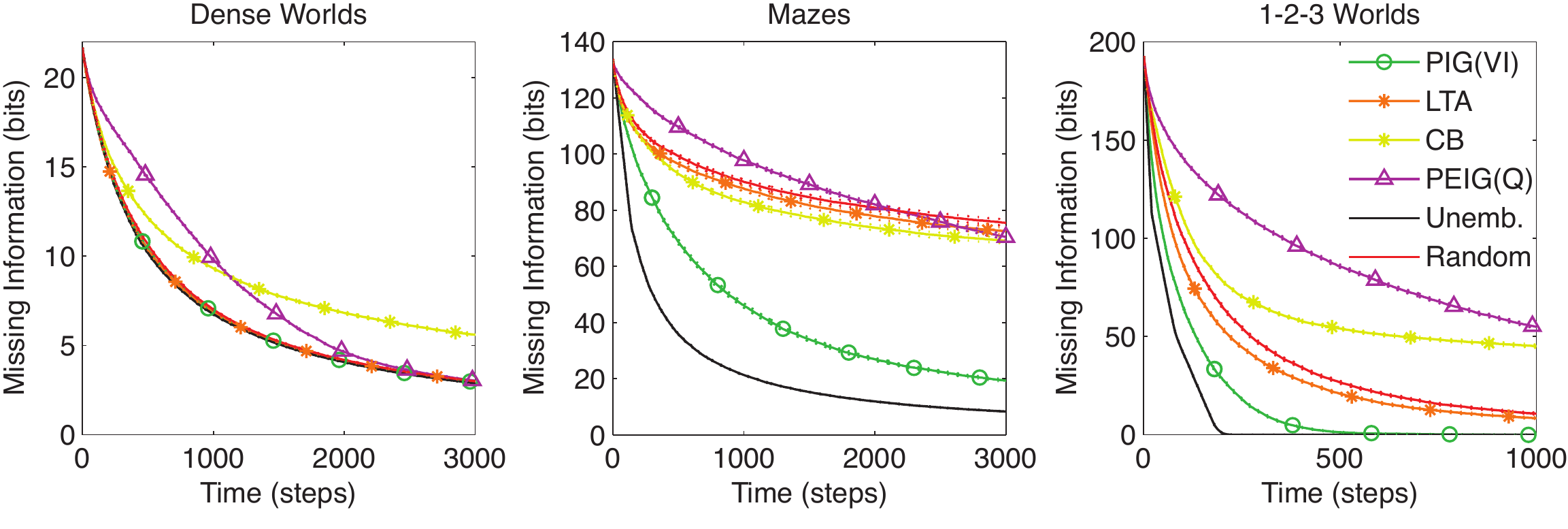}
\caption{Comparison to previous exploration strategies. The average missing information is plotted over time for PIG(VI) agents along with three exploration strategies from the literature: least taken action (LTA) \cite{si2007gain,barto1990computational,sato1988learning}, counter-based (CB)  \cite{thrun1992}, and Q-Learning on posterior expected information gain (PEIG(Q))  \cite{storck1995}. The standard control strategies are also shown. Standard errors are plotted as dotted lines above and below learning curves. (n=200)}
\label{fig:Comp}
\end{figure*}

While exploration in the RL literature has largely focused on its role in reward acquisition, many of the principles developed to induce exploration can be implemented in our framework. In this section, we compare these various methods to PIG(VI) under our learning objective.

Random action is perhaps the most common exploration strategy used in RL. As we have already seen in Fig. \ref{fig:Coord}, random action is only efficient for exploring \rntwks{}. In addition to undirected random action, the following directed exploration strategies have been developed in the RL literature. The learning curves of the various strategies are plotted in Fig. \ref{fig:Comp}.

{\it Least Taken Action (LTA):} Under LTA, an agent will always choose the action that has been performed least often in the current state \cite{si2007gain,barto1990computational,sato1988learning}. Like random action, LTA yields uniform sampling of actions in each state. Consistently, LTA fails to significantly improve on the learning rates seen under random action ($p>0.001$ for all three environments).

{\it Counter-Based Exploration (CB):} Whereas LTA actively samples actions uniformly, CB attempts to induce a uniform sampling across states. To do this, it maintains a count of the occurrences of each state, and chooses its action to minimize the expected count of the resultant state \cite{thrun1992}. As shown in Fig. \ref{fig:Comp}, CB performs even worse than random action in \rntwks{} and \sntwks{} ($p<0.001$). It does outperform random actions in Mazes but falls far short of the performance seen by PIG(VI) ($p<0.001$). 

{\it Q-learning on Posterior Expected Information Gain (PEIG(Q)):} Stork {\it et al.} \cite{storck1995} developed a utility function $U_{Storck}$ to measure past changes in the internal model, which they used to guide exploration under a Q-learning algorithm \cite{sutton1998}. Let $\tau$ be the most recent time step in the past over which the internal model for the transition distribution $\bT_{a,s,:}$ changed:
\begin{equation*}  
\tau:=\max\{t|s(t)=s,a(t)=a,t<|\D|\}
\end{equation*}
Then, considering the internal model before and after this time step ($\bTh^{\tau}$ and $\bTh^{\tau+1}$ respectively), and the data collected up to this point $\D^{\tau+1}$, the utility function defined by Storck {\it et al.} is:
\begin{equation}
U_{Storck}:=\KL[\bTh^{\tau+1}_{a,s,:}\parallel \bTh^{\tau}_{a,s,:}]
\end{equation}
Note, both $\bTh^{\tau}$ and $\bTh^{\tau+1}$ are internal models previously (or currently) held by the agent. In the following derivation, we demonstrate that $U_{Storck}$ is equivalent to the posterior expected information gained (PEIG).
\begin{IEEEeqnarray*}{rCl}
\PEIG(a,s)&:=&\E_{\bT|\D^{\tau+1}}\left[\IM(\bT\parallel\bTh^{\tau})-\IM(\bT\parallel\bTh^{\tau+1})\right]\\
&=&\E_{\bT|\D^{\tau+1}}\left[\sum_{s'}\bT_{a,s,s'}\log_2\left(\frac{\bTh^{\tau+1}_{a,s,s'}}{\bTh^{\tau}_{a,s,s'}}\right)\right]\\
&=&\sum_{s'}\E_{\bT|\D^{\tau+1}}[\bT_{a,s,s'}]\log_2\left(\frac{\bTh^{\tau+1}_{a,s,s'}}{\bTh^{\tau}_{a,s,s'}}\right)\\
&=&\sum_{s'} \bTh^{\tau+1}_{a,s\ra s'} \log_2\left(\frac{\bTh^{\tau+1}_{a,s\ra s'}}{\bTh^{\tau}_{a,s\ra s'}}\right)\\
&=&\KL[\bTh^{\tau+1}_{a,s,:}\parallel \bTh^{\tau}_{a,s,:}] \label{eq:PEIG}\IEEEyesnumber%
\end{IEEEeqnarray*} 
Thus, PEIG is a posterior analogue to our PIG utility function. Q-learning is a model-free approach to maximizing long-term gains of a utility function \cite{sutton1998}. Following Storck {\it et al.}, we tested the combination of PEIG and Q-learning (PEIG(Q)) in our test environments. Surprisingly, PEIG(Q) performs even worse, at least initially, than random action in all three environments ($p<0.001$ for CMCs and \sntwks{}, $p>0.001$ for Mazes). As such, it fails to yield the learning performance seen by PIG(VI) in Mazes in \sntwks{}.

Altogether, Fig. \ref{fig:Comp} demonstrates that PIG(VI) outperforms the previous explorative strategies at learning structured worlds. To further compare the principles of PIG(VI) and PEIG(Q), we introduce two cross-over strategies that borrow from each of them. The first is PIG(Q) which applies Q-learning to the PIG utility function. The learning performance of PIG(Q) is similar to PEIG(Q), falling short of PIG(VI) (Fig. S3). This suggests that Q-learning is ineffective at coordinating actions during exploration. The second cross-over strategy is PEIG(VI) which applies the VI algorithm to Storck {\it et al.}'s utility function. PEIG(VI) matched PIG(VI) in Mazes ($p>0.001$) but not \sntwks{} ($p<0.001$), suggesting that the posterior information gain is a reasonable predictor for future information gain under a Dirichlet prior but not a Discrete prior.

\begin{figure*}[!tH]
\centering
\includegraphics[scale=0.7]{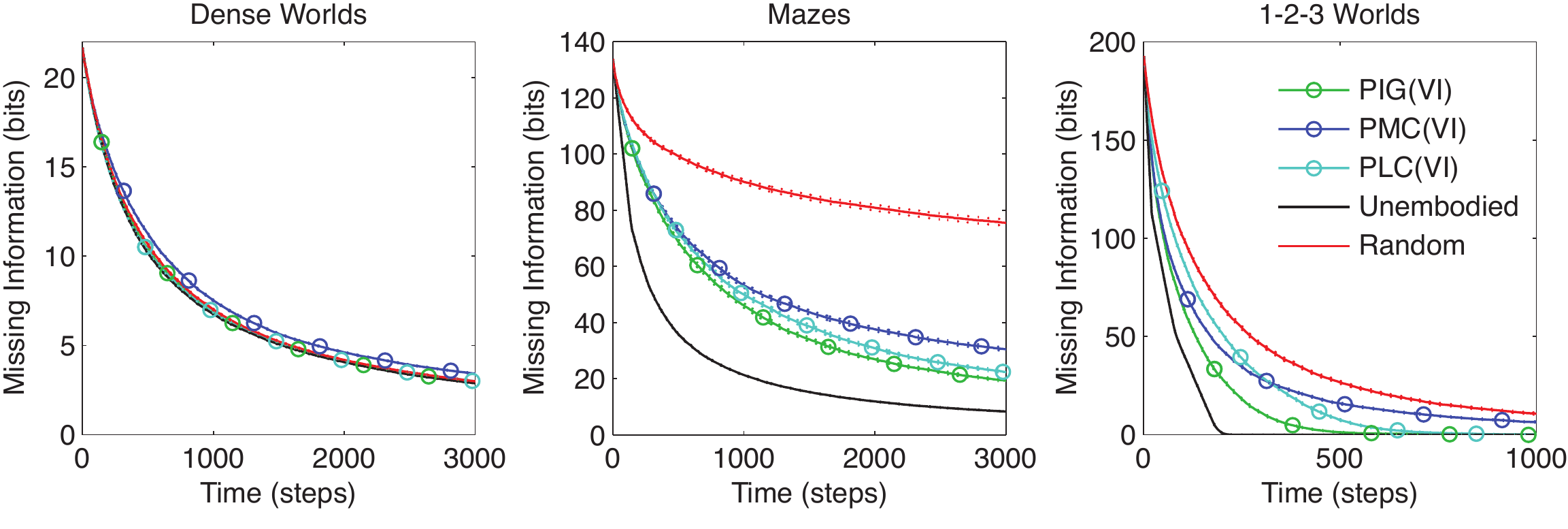}
\caption{Comparison between utility functions. The average missing information is plotted over time for agents that employ VI  to maximize long-term gains in the three objective function, PIG, PMC, or PLC. The standard control strategies are also shown. Standard errors are plotted as dotted lines above and below learning curves. (n=200)}
\label{fig:Psych}
\end{figure*}

\subsection{Comparison to utility functions from Psychology}
Inspired by independent findings in the field of Psychology that PIG can describe human behavior during hypothesis testing, we investigated two other measures also developed in this context \cite{oaksford1994rational,nelson2005}. Like PIG, both are measures of the difference between the current and hypothetical future internal models:

{\it Predicted mode change (PMC)} predicts the height difference between the modes of the current and future internal models \cite{nelson2005,baron1985rationality}:  
\begin{equation}
\PMC(a,s)=\sum_{s^*}\bTh_{s,a\ra s^*}\left[\max_{s'}\bTh^{a,s,s^*}_{a,s,s'}-\max_{s'}\bTh_{a,s,s'}\right]\label{eq:PMC}
\end{equation} 

{\it Predicted L1 change (PLC)} predicts the average L1 distance between the current and future internal models \cite{klayman1987confirmation}:
\begin{equation}
\PLC(a,s)=\sum_{s^*}\bTh_{s,a\ra s^*}\left[\frac{1}{N}\sum_{s'}\left|\bTh^{a,s,s^*}_{a,s\ra s'}-\bTh_{a,s\ra s'}\right|\right]\label{eq:PLC}
\end{equation} 

Note, PMC and PLC differ from PIG in the norm used to quantify differences between CMC kernels. Considering an arbitrary norm $d$, the claim analogous to Theorem 2 would be:
\begin{IEEEeqnarray*}{l}
\sum_{s^*}\bTh^{current}_{a,s,s^*}d(\bTh^{future}\parallel\bTh^{current})\\
\;\;\;\;\;\;\;\;\;\;\;\;\;=\E_{s^*,\bT|\D}\left[d(\bT\parallel\bTh^{current})-d(\bT\parallel \bTh^{future})\right]
\end{IEEEeqnarray*}
This claim states that the expected difference between the current and future internal model equals the expected change in difference with respect to the ground truth.  While this claim holds when $d$ is the norm used in PIG (Theorem 2), it does not generally hold for either of the norms used in PMC or PLC.

To our knowledge, neither PIG, PMC nor PLC have previously been applied to sequences of observations or to embodied action perception loops. We tested agents that attempt to maximize PMC or PLC using VI. As Fig. \ref{fig:Psych} reveals, PIG(VI) proved again to be the best performer overall. In particular, PIG(VI) significantly outperforms PMC(VI) in all three environments, and PLC(VI) in \sntwks{} ($p<0.001$). Nevertheless, PMC and PLC achieved significant improvements over the baseline control in Mazes and \sntwks{}, highlighting the benefit of value iteration across different utility functions. 
Interestingly, when performance was measured by an L1 distance instead of missing information, PIG(VI) still outperformed PMC(VI) and PLC(VI) in \sntwks{} (Fig. S4). 

\subsection{Generalized utility of exploration}
From a behavioral perspective, learning represents a fundamental and primary drive \cite{loewenstein1994psychology,archer1983exploration}. The evolutionary advantage of such an exploratory drive likely rests on the general utility of the acquired internal model \cite{kaplan1983cognition,pisula2003costs,pisula2008play,Renner_1988,renner1990neglected}. To test this, we assessed the ability of the agents to use their internal models, derived through exploration, to accomplish an array of goal-directed tasks. We consider two groups of tasks: navigation and reward acquisition. 

{\it Navigation:} Starting at any given state, the agent has to navigate to any given target state with the minimal number of steps.

{\it Reward Acquisition:} For every starting state, the agent has to gather as much reward as possible over 100 time steps. Reward values are drawn from a standard normal distribution and randomly assigned to every state in the CMC. Each agent is tested in ten randomly generated reward structures.

At several time points during exploration, the agent is stopped and its internal models assessed for general utility. For each task, we next derive the behavioral policy that optimizes performance under the internal model. The derived policy is then tested in the world (i.e. under the true CMC kernel), and the expected path length or acquired reward for that policy is determined. As a positive control, we also derive an objective optimal policy that maximizes the realized performance for the true CMC kernel. The difference in realized performance between the subjective and objective policies is used as a measure of navigational loss or reward loss. High navigational loss means the agents policy took many more time steps to reach the target state than the optimal policy. High reward loss means the agents policy yielded significantly fewer rewards than the optimal policy.

\begin{figure*}[!t]
\centering
\includegraphics[scale=0.6]{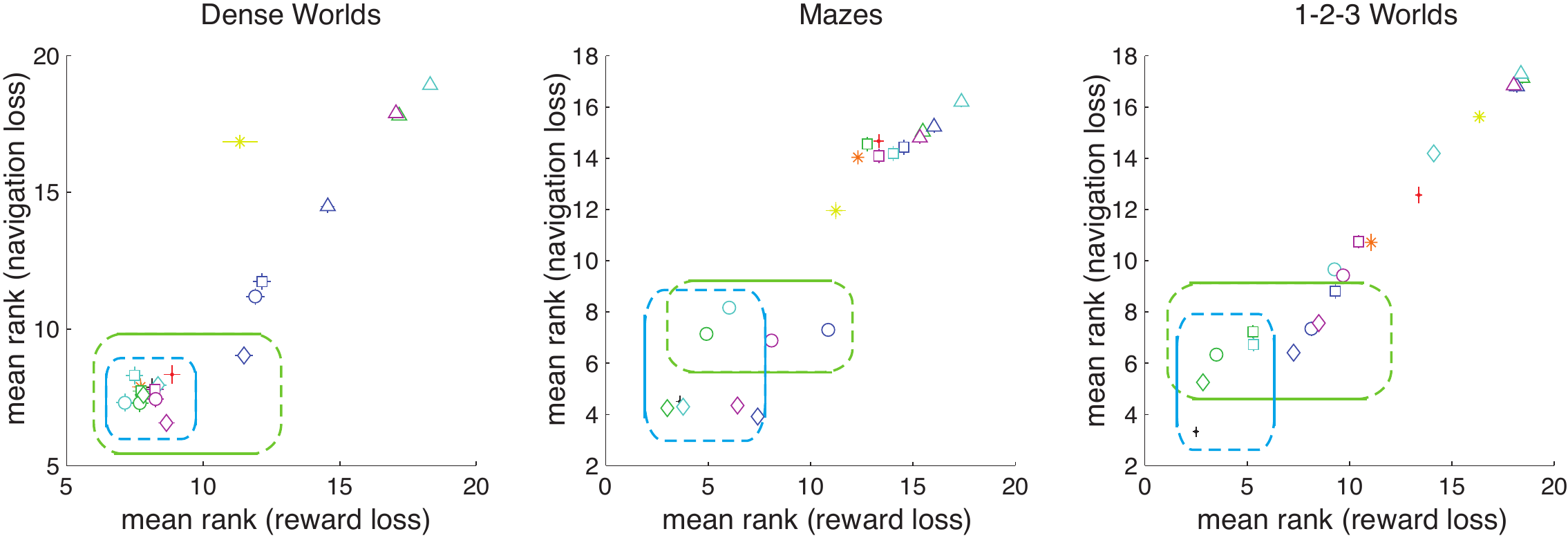}
\caption{Demonstration of generalized utility. For each world (n=200), explorative strategies are ranked for average navigational loss (averaged across N start states and N target states) and average reward loss (averaged across N start states and 10 randomly generated reward distributions). The average ranks are plotted with standard deviations. Strategies lying outside the pair of solid green lines differ significantly from PIG(VI) in navigational loss. Strategies lying outside the pair of solid blue lines differ significantly from PIG(VI) in reward loss ($p<0.0001$). The different utility functions and heuristics are distinguished by color: PIG(green), PEIG(magenta), PMC(blue), PLC(cyan), LTA(orange), CB(yellow). The different coordination methods are distinguished by symbol: Greedy(squares), VI(circles), VI+(diamonds), Heuristic Strategies(asterisks). The two standard controls are depicted as follows: Unembodied(black), Random(red).}
\label{fig:navrew}
\end{figure*}

Fig. \ref{fig:navrew} depicts the average rank in navigational and reward loss for the different explorative strategy. Significance bounds ($p=0.001$) around PIG(VI) were determined by post-hoc analysis of Friedman's test \cite{hochberg1987multiple}. In all environments, for both navigation and reward acquisition, PIG(VI) always grouped with the top performers ($p>0.001$), excepting positive controls. PIG(VI) was the only strategy to do so. Thus, the explorative strategy that optimized learning under the missing information objective function gave the agent an advantage in a range of independent tasks. 

\section{Discussion}

In this manuscript we introduced a parsimonious mathematical framework for studying learning-driven exploration by embodied agents based on information theory, Bayesian inference and controllable Markov chains (CMCs). We compared agents that utilized different exploration strategies towards optimizing learning. To understand how learning performance depends on the structure of the world, three classes of environments were considered that challenge the learning agent in different ways. We found that fast learning could be achieved by an exploration strategy that coordinated actions towards long-term maximization of predicted information gain (PIG).

\subsection{Potential limitations to our approach}

The optimality of the Bayesian Estimate (Theorem 1) and the accuracy of PIG (Theorem 2) both require a prior distribution on the transition distributions. For biological agents, such priors could have been learned from earlier exploration of related environments, or may represent hardwired beliefs optimized by evolutionary pressures. As another possibility, an agent could attempt to simultaneously learn a prior while exploring its environment. Indeed, additional results (Fig. S5) show that the maximum-likelihood estimation of the concentration parameter for \rntwks{} and Mazes enables exploration that quickly matches the performance of agents given accurate priors. Nevertheless, biological agents may not always have access to an accurate prior for an environment. For such cases, future work is required to understand exploration under false priors and how they could yield sub-optimal but perhaps biologically realistic exploratory behaviors.

As another potential limitation of our approach, the VI algorithm is only optimal for dynamic processes with known stationary transition probabilities and stationary utilities \cite{Bel57}.  In contrast, any utility function, including PIG, that attempts to capture the progress in learning of an agent will necessarily change over time. This caveat may be partially alleviated by the fact that PIG changes only for the sampled distributions. Furthermore, PIG decreases in a monotonic fashion (see Fig. \ref{fig:PIGvsIG}) which could potentially be captured by the discount factor of VI. Interesting future work may lie in accounting for the effect of such monotonic decreases in estimates of future learning gains.

In addition, the learning agent does not have access to the true transition distributions for performing VI and has to rely instead on its evolving internal model. The impairment caused by this reliance on the internal model was directly  assessed with a positive control PIG(VI+). A comparison of PIG(VI) against this control (Fig. \ref{fig:Coord}) shows performance impairment only in Mazes and it is rather small compared to the improvements offered by VI.

Finally, it might be argued that the use of missing information as a measure of learning unfairly advantaged the PIG utility function. Interestingly, however, PIG under VI was not only the fastest learner, but also demonstrated the greatest capacity for accomplishing goal-directed tasks. Furthermore, it even outperformed other strategies, including PLC(VI), under an L1 objective function (Fig. S4).

\subsection{Related work in Reinforcement Learning}

CMCs are closely related to the more commonly studied Markov Decision Processes (MDPs) used in Reinforcement Learning. MDPs differ from CMCs in that they explicitly include a stationary reward function associated with each transition \cite{sutton1998,gimbert2007pure}. RL research of exploration usually focusses on its role in balancing exploitative behaviors during reward maximization. Several methods for inducing exploratory behavior in RL agents have been developed. Heuristic strategies such as random action, least taken action, and counter-based algorithms are commonly employed in the RL literature. While such strategies may be useful in RL, our results show that they are inefficient for learning the dynamics of structured worlds.

In contrast to these heuristic strategies for exploration, several principled approaches have been proposed for inducing exploratory actions to maximize rewards. For example, the BEETLE algorithm models reward as a partially observable MDP and derives an analytic solution to optimize rewards \cite{poupart2006analytic}. Similarly, the BOSS approach maintains a posterior distribution over MDPs from which it periodically samples for selecting actions that maximize reward gains "optimistically" from the samples \cite{asmuth2009bayesian}. These strategies focus exclusively on extrinsically motivated exploration and do not address exploration driven by learning for its own sake.

Finally, several studies have investigated intrinsically motivated learning under the RL framework. For example, Singh et al. \cite{singh2010intrinsically} have demonstrated that RL guided by saliency, an intrinsic motivation derived from changes in stimulus intensity, can promote the learning of reusable skills. As mentioned previously, Storck et al. introduced the combination of Q-learning and PEIG as an intrinsic motivator of learning \cite{storck1995}. In their study, PEIG(Q) outperformed random action only over long time scales. At shorter time scales, random action performed better. Interestingly, we found exactly the same trend, initially slow learning with eventual catching-up, when we applied PEIG(Q) to exploration in our test environments (Fig.~6).

\subsection{Related work in Psychology}
In the Psychology literature, PIG, as well as PMC and PLC, were directly introduced as measures of the expected difference between a current and future belief \cite{oaksford1994rational,nelson2005,baron1985rationality,klayman1987confirmation}. Here, in contrast, we derived PIG, using Bayesian inference, from the expected change in missing information with respect to a ground truth (Theorem 2). Analogous theorems do not hold for PMC or PLC. For example, the expected change in L1 distance between an internal model and the true structure is not equivalent to the expected L1 distance between successive internal models. This might explain why PIG(VI) outperformed PLC(VI) even under an L1 measure of learning (Fig. S4). 

We applied the PIG principle to the learning of a full model of the world. The Psychology literature, in contrast, focusses on specific questions (hypothesis testing) regarding the data. In addition, this prior literature has not considered sequences of actions or embodied sensor-motor loops.

It has been shown that human behavior during hypothesis testing can be explained by a model that maximizes PIG \cite{oaksford1994rational,nelson2005}. This suggests that the PIG information-theoretic measure may have biological significance. The behavioral studies, however, could not distinguish between the different utility functions (PIG, PMC and PLC) in their ability to explain human behavior \cite{nelson2005}. Perhaps our finding that \sntwks{} give rise to large differences between the three utility functions can help identify new behavioral tasks for disambiguating the role of these measures in human behavior.

Itti and Baldi recently developed an information theoretic measure closely related to PEIG for modeling bottom-up visual saliency and predicting gaze attention \cite{itti2006,Baldi_Itti10nn,Itti_Baldi09vr}. In their model, a Baysian learner maintains a probabilistic belief structure over the low-level features of a video. Attention is believed to be attracted to locations in the visual scene that exhibit high Surprise. Like PEIG, Surprise quantifies changes in posterior beliefs by a summed Kullback-Leibler divergence. Several potential extensions of this work are suggested by our results. First, it may be useful to model the active nature of data acquisition during visual scene analysis. In Itti and Baldi's model, all features are updated for all location of the visual scene regardless of current gaze location or gaze trajectory. Differences in accuity between the fovea and periphery however suggest that gaze location will have a significant effect on which low-level features can be transmitted by the retina \cite{wassle1991functional}. Second, our comparison between the PIG and PEIG utility functions (Figs. 6 and S3) suggests that predicting where future change might occur, may be more efficient than focusing attention only on those locations where change has occured in the past. A model that anticipates Surprise, as PIG anticipates information gain, may be better able to explain some aspects of human attention. For example, if a moving subject disappears behind an obscuring object, viewers may anticipate the reemergence of the subject and attend the far edge of the obscurer. Incorporating these insights into new models of visual saliency and attention could be an interesting course of future research.

\subsection{Information-theoretic models of behavior}
The field of behavioral modeling has recently seen increased utilization of information-theoretic concepts. These approaches can be grouped under three guiding principles. The first group uses information theory to quantify the complexity of a behavioral policy, with high complexity generally considered undesirable. Tishby and Polani for example, considered RL maximization of rewards under such complexity constraints \cite{tishby2011information}. While we did not consider complexity constraints on our behavioral strategies in the current work, it may be an interesting topic for future studies.

The second common principle seeks to maximize {\it predictive information} \cite{tishby2000information, ayetal2008,still2009} (not to be confused with predicted information gain, PIG). Predictive information, which has also been termed {\it excess entropy} \cite{crutchfield2003regularities}, estimates the amount of information a known variable (or past variable) contains regarding an unknown (or future) variable. For example, in simulated robots, Ay et al. demonstrated that complex and interesting behaviors can emerge by choosing control parameters that maximize the predictive information between successive sensory inputs \cite{ayetal2008}. The information bottleneck approach introduced by Tishby et al. \cite{tishby2000information} combines predictive information and complexity constraints, maximizing the information between a compressed internal variable and future state progression subject to a constraint on the complexity of generating the internal variable from sensory inputs. Recently, Still extended the information bottleneck method to incorporate actions \cite{still2009}. 

Both Ay {\it et al.} and Still describe the behaviors that result from their models as exploratory. Their objective of high predictive information selects actions such that the resulting sensory input changes often but in a predictable way. We therefore call this form of exploration stimulation-driven. Predictive information can only be high when the sensory feedback can be predicted, and thus stimulation-driven exploration relies on an accurate internal model. In contrast, the learning objective we introduced here drives actions most strongly when the internal model can be improved and this drive weakens as it becomes more accurate.  Thus, learning-driven and stimulation-driven exploration contrast each other while being very interdependent. Indeed, a simple additive combination of the two objectives may naturally lead to a smooth transitioning between the two types of exploration, directed by the expected accuracy of the internal model. In the next section we suggest a correspondence of these two computational principles of exploration with two distinct modes of behavior distinguished in psychology and behavioral research.

Finally, the Free-Energy (FE) hypothesis introduced by Friston proposes that the minimization of free-energy, an information-theoretic bound on surprise, offers a unified variational principle for governing both the learning of an internal model as well as actions \cite{friston2009free}. Friston notes that under this principle agents should act to minimize the number of states they visit. This stands in stark contrast to both learning-driven and stimulation-driven exploration. During learning-driven exploration, an agent will seek out novel states where missing information is high. During stimulation-driven exploration, an agent will actively seek to maintain high variation in its sensory inputs. Nevertheless, as Friston argues, reduced state entropy may be valuable in dangerous environments where few states permit survival. The balance between cautionary and exploratory behaviors would be an interesting topic for future research.

\subsection{Towards a general theory of exploration}
With the work of Berlyne \cite{berlyne1966curiosity}, Psychologists began to dissect the complex domains of behavior and motivation that comprise exploration. A distinction between play (or diversive exploration) and investigation (or specific exploration) grew out of two competing theories of exploration. As reviewed by Hutt \cite{hutt1970specific}, "curiosity"-theory proposed that exploration is a consummatory response to curiosity-inducing stimuli \cite{berlyne1950novelty, montgomery1953exploratory}. In contrast, "boredom"-theory held that exploration was an instrumental response for stimulus change \cite{glanzer1958curiosity,myers1954failure}. To ameliorate this opposition, Hutt suggested that the two theories may be capturing distinct behavioral modes, with "curiosity"-theory underlying investigatory exploration and "boredom"-theory underlying play. In children, exploration often occurs in two stages, inspection to understand what is perceived, followed by play to maintain changing stimulation \cite{hutt1972predictions}. These distinctions nicely correspond to the differences between our approach and the predictive information approach of Ay {\it et al.} \cite{ayetal2008} and Still \cite{still2009}. In particular, we hypothesize that our approach, which emphasizes the acquisition of information, corresponds to curiosity-driven investigation. In contrast, we propose that predictive information a la Ay {\it et al.} and Still, which rehearses the internal model in a wide range, may correspond with play. Further, the proposed method of additively combining these two principles (Section 4.4), may naturally capture the transition between investigation and play seen in children during exploration. 

Even in the domain of curiosity-driven exploration, there are many varied theories \cite{loewenstein1994psychology}. Early theories viewed curiosity as a drive to maintain a specific level of arousal. These were followed by theories interpreting curiosity as a response to intermediate levels of incongruence between expectations and perceptions, and later by theories interpreting curiosity as a motivation to master one's environment. Loewenstein developed an Information Gap Theory and suggested that curiosity is an aversive reaction to missing information \cite{loewenstein1994psychology}. More recently, Silvia proposed that curiosity is composed of two appraisal components, complexity and comprehensibility. For Silvia complexity is broadly defined, and includes novelty, ambiguity, obscurity, mystery, etc. Comprehensibility appraises whether something can be understood. It is interesting how well these two appraisals match information-theoretic concepts, complexity being captured by entropy, and comprehensibility by information gain \cite{pfaffelhuber1972}. Indeed, predicted information gain might be able to explain the dual appraisals of curiosity-driven exploration proposed by Silvia. PIG is bounded by entropy and thus high values require high complexity. At the same time, PIG equals the expected decrease in missing information and thus may be equivalent to expected comprehensibility.

All told, our results add to a bigger picture of exploration in which the theories for its different aspects fit together like pieces of a puzzle. This invites future work for integrating these pieces into a more comprehensive theory of exploration and ultimately of autonomous behavior.

\section*{Acknowledgements}
The authors wish to thank Nihat Ay, Susanne Still,  Jim Crutchfield, Reza Moazzezi, and the Redwood Center for Theoretical Neuroscience for useful discussions. D.Y. Little is supported by the Redwood Center for Theoretical and Computational Neuroscience Graduate Student Endowment. This work was supported in part by the National Science Foundation grant CNS-0855272.

\bibliographystyle{IEEEtranS}
\bibliography{IEEEabrv,EmbodiedExploration}

\newpage
\begin{figure*}[!tH]
\centering
\includegraphics[scale=1]{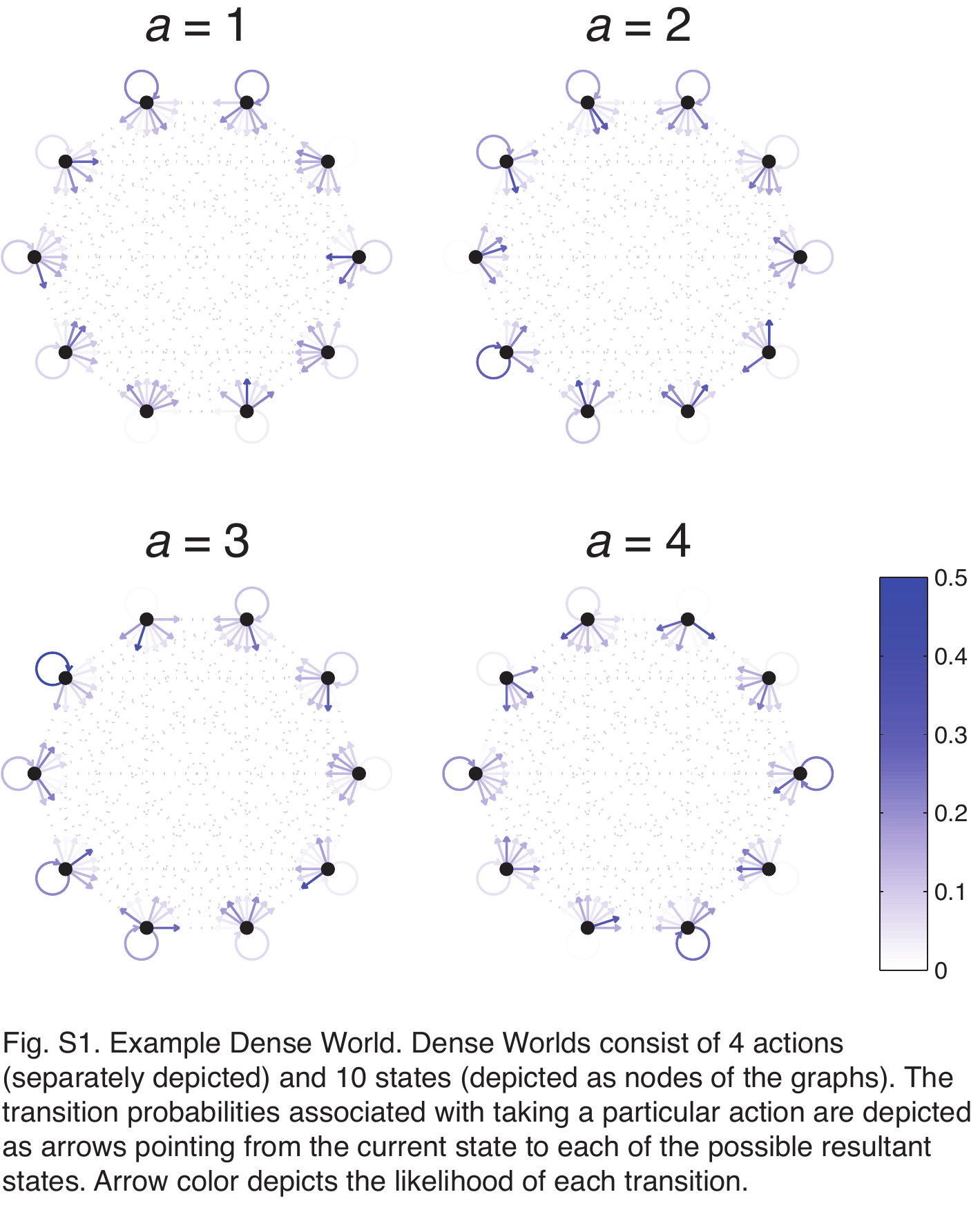}
\end{figure*}

\newpage
\begin{figure*}[!tH]
\centering
\includegraphics[scale=1]{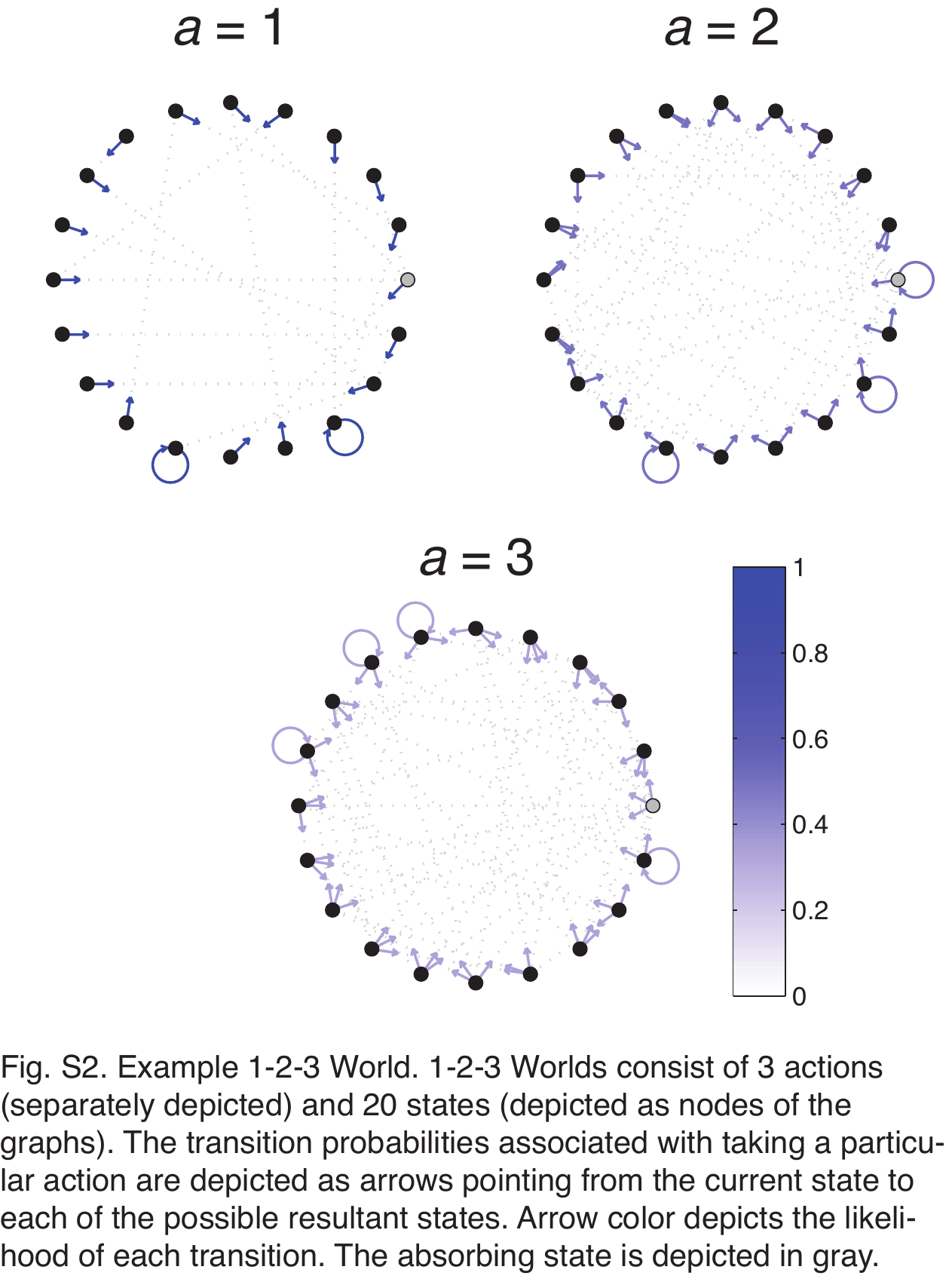}
\end{figure*}

\newpage
\begin{figure*}[!H]
\centering
\includegraphics[scale=0.7]{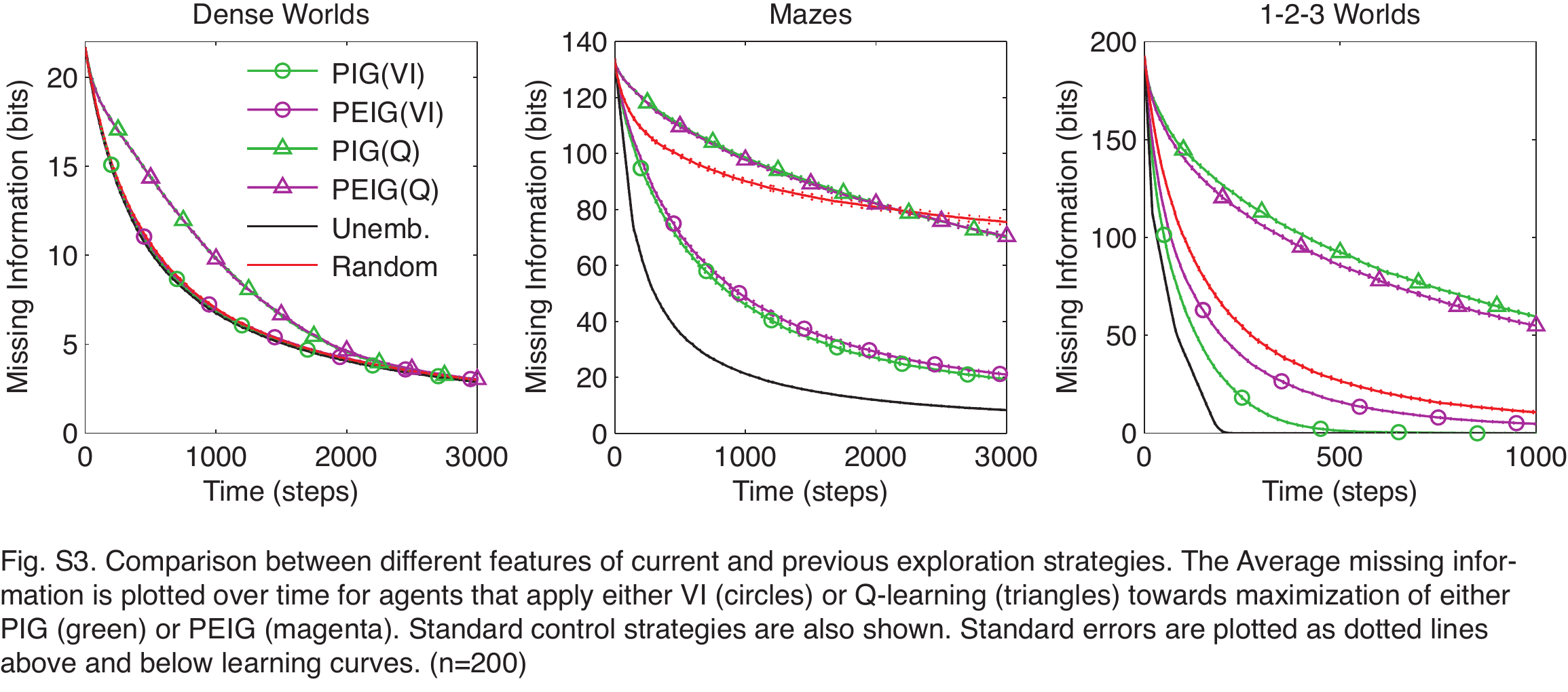}
\end{figure*}

\newpage
\begin{figure*}[!H]
\centering
\includegraphics[scale=0.7]{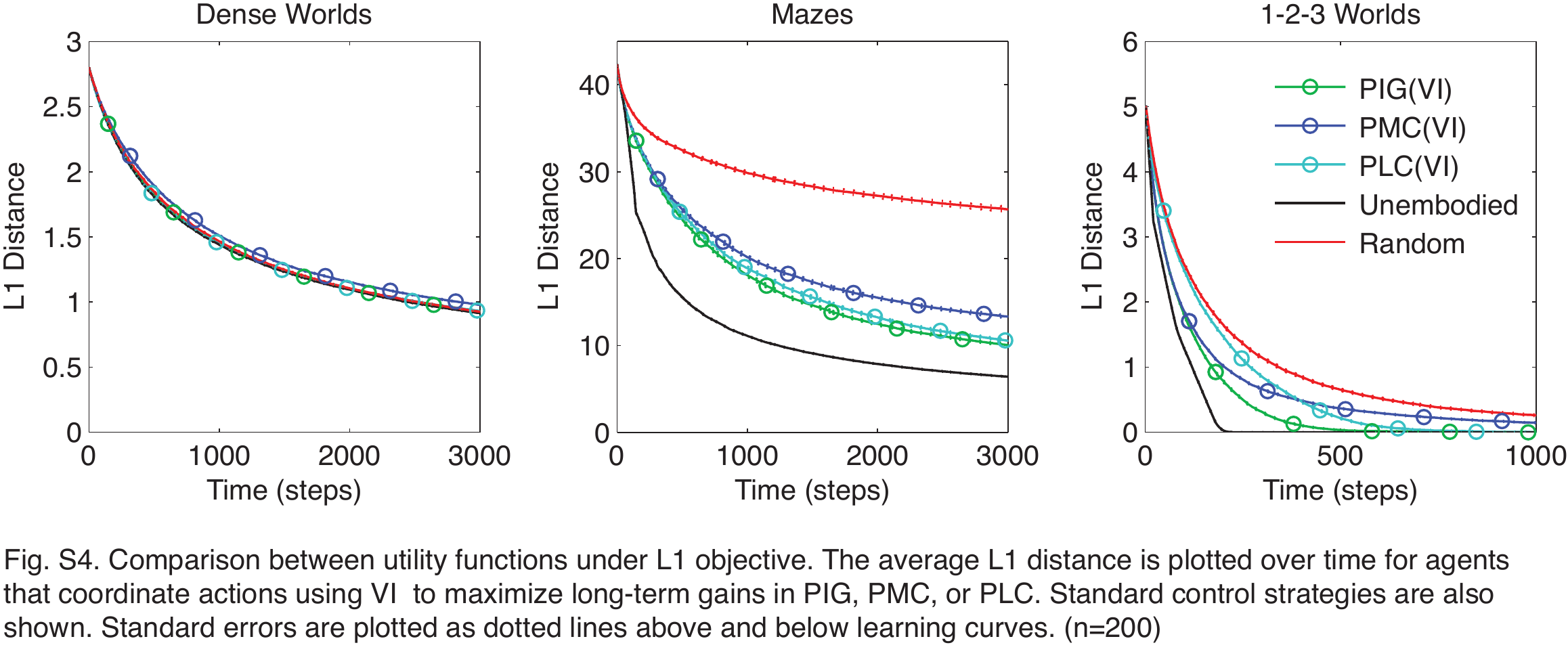}
\end{figure*}

\newpage
\begin{figure*}[!tH]
\centering
\includegraphics[scale=0.7]{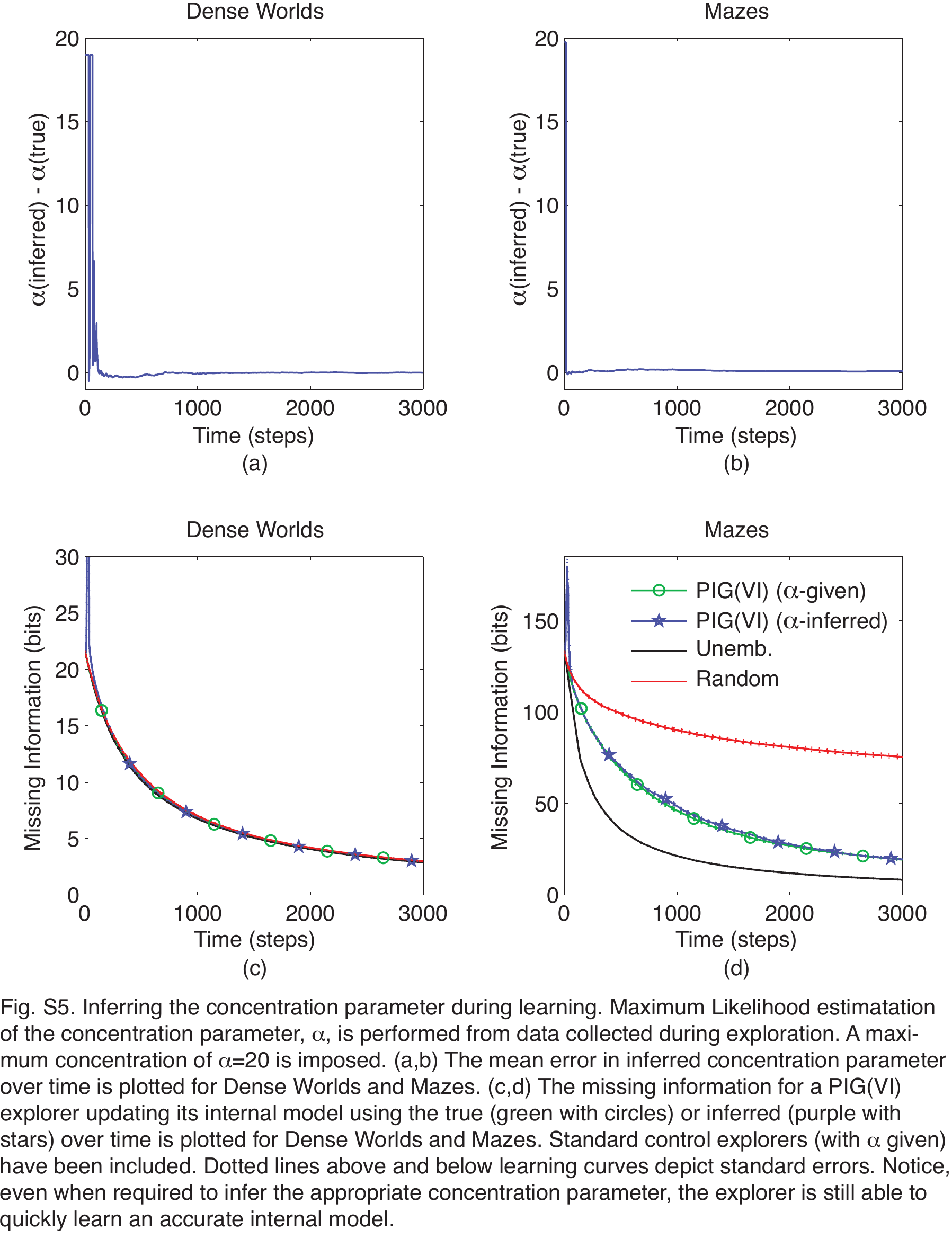}
\end{figure*}
%
%

\end{document}